\documentclass[journal,12pt,onecolumn,doublespacing]{IEEEtran}
\usepackage{url}
\usepackage{graphicx}
\usepackage{amsfonts}
\usepackage{amssymb}
\usepackage{amsthm}
\usepackage{color}
\usepackage{amsmath}
\usepackage{caption}
\usepackage[mathscr]{eucal}
\usepackage{multirow}
\usepackage{algorithm}
\usepackage{algorithmic}
\usepackage{epstopdf}
\usepackage{float}
\usepackage{subfigure}

\newcommand{\R}{\mathbb{R}}
\newcommand{\C}{\mathbb{C}}
\newcommand{\diag}{\text{Diag}}
\newcommand{\tr}{\text{tr}}
\newcommand{\iFFT}[1]{\mathcal{F}^{-1} ( #1 )} 
\newcommand{\FFT }[1]{\mathcal{F}( #1 )} 

\newcommand{\NM}[2]{\left\|  #1 \right\|_{#2}}
\newcommand{\HH}[1]{\mathcal{H}( #1 )}

\newtheorem{theorem}{Theorem}
\newtheorem{lemma}[theorem]{Lemma}
\newtheorem{proposition}[theorem]{Proposition}
\newtheorem{assumption}{Assumption}

\begin{document}
\title{Scalable Online Convolutional Sparse Coding}

\author
{
	Yaqing~Wang,
	Quanming~Yao,	
	James~T.~Kwok,
	and~Lionel~M.~Ni.
	\thanks{Y. Wang, Q. Yao and J. T. Kwok are with Department of Computer Science and Engineering, Hong Kong University of Science and Technology University, Hong Kong.}
	\thanks{L.M. Ni is with Department of Computer and Information Science,
		University of Macau, Macau.}
}

\markboth{}%
{}

\maketitle

\begin{abstract}
Convolutional sparse coding (CSC) improves sparse coding by learning a shift-invariant
dictionary
from the data. However, existing CSC algorithms operate in the batch mode and are expensive,
in terms of both space and time, on large data sets. In this paper, 
we alleviate these problems by
using online learning.
	The key is a reformulation of the CSC objective
so that 
convolution 
can be handled easily 
in the frequency domain 
and much smaller history matrices are needed.
We use 
the alternating direction method of multipliers (ADMM) 
to solve the resultant optimization problem, and the 
ADMM subproblems have efficient closed-form solutions.
Theoretical analysis shows that the learned dictionary converges to a stationary point of the optimization problem.
Extensive experiments show that
convergence of the proposed method is much faster and its reconstruction performance is also better.
Moreover, while existing CSC algorithms can only run on a small number of images,
the proposed method can handle at least 
ten times more images.
\end{abstract}


\section{Introduction}
\label{sec:intro}

\IEEEPARstart{I}{n} 
recent years,
sparse coding has been widely used in signal processing \cite{aharon2006rm,lee2007efficient} and computer vision \cite{mairal2009non,yang2009linear}.
In sparse coding,
each data sample is represented as a weighted combination of a few atoms from an 
over-complete dictionary learned from the data.
Despite its popularity, sparse coding cannot capture shifted local patterns that are common
in image samples.
Often, it has to first extract overlapping image patches,
which is analogous to manually convolving
the dictionary with the samples.
As each sample element (e.g., an image pixel) is contained in multiple
overlapping patches, the separately learned representations may not be consistent. Moreover, the resultant representation is highly redundant \cite{bristow2013fast}.

Convolutional sparse coding (CSC) addresses this problem by
learning a shift-invariant dictionary composed of many filters.
Local patterns at translated positions of the samples are easily extracted by 
convolution, and eliminates the need for generating overlapping patches. 
Each sample is approximated by the sum of a set of filters convolved with the corresponding
codes. The learned representations are 
consistent as they are obtained together.
CSC has been used successfully in various image processing applications 
such as super-resolution image reconstruction \cite{gu2015convolutional}, high dynamic range imaging \cite{serrano2016convolutional}, image denoising and inpainting \cite{heide2015fast}.
It is also popular in biomedical applications, 
e.g., cell identification \cite{pachitariu2013extracting}, calcium image analysis \cite{andilla2014sparse}, tissue histology classification \cite{chang2017unsupervised} and 
segmentation 
of curvilinear structures 
\cite{annunziata2016accelerating}. 
CSC has also been used in audio processing applications such as piano music transcription \cite{cogliati2016context}.

A number of approaches have been proposed to 
solve the optimization problem in CSC. 
In the pioneering 
{\em deconvolutional network} (DeconvNet) \cite{zeiler2010deconvolutional},
simple gradient descent
is used.
As convolution is slow in the spatial domain,
{\em fast convolutional sparse coding} (FCSC) \cite{bristow2013fast} formulates CSC in the
frequency domain, and
the alternating direction method of multipliers (ADMM) \cite{boyd2011distributed} 
is used to solve the resultant optimization problem.   
Its most expensive operation is the inversion of a convolution-related linear operator.  To
alleviate this problem,
{\em convolutional basis pursuit denoising} (CBPDN) \cite{wohlberg2016efficient}
exploits a special structure of the dictionary, while
the {\em global consensus ADMM} (CONSENSUS) \cite{sorel2016fast}
utilizes the matrix inverse lemma to simplify computations.
{\em Fast and flexible convolutional sparse coding}
(FFCSC) 
\cite{heide2015fast} further introduces mask matrices so as to handle incomplete samples
that  are common in 
image/video inpainting and demosaicking
applications.
Note that all these algorithms operate in the batch mode (i.e.,
all the samples/codes have to be accessed in each iteration).  Hence they 
can become expensive, in terms of both space and time, on large data sets.



In general,
online learning has been commonly  used 
to improve the scalability of machine learning algorithms 
\cite{cesa-bianchi-06,shalev2012online}.
While batch learning algorithms train the model after arrival of
the whole data set,
online learning algorithms observe the samples sequentially and update the model incrementally.
Moreover, data samples need not be stored after being processed. 
This can significantly reduce the algorithm's time and space complexities.
In the context of sparse coding, an efficient online algorithm
is proposed in \cite{mairal2010online}.  
In each iteration, 
information necessary for dictionary update is summarized in 
fixed-sized history matrices.
The space complexity of the algorithm is thus independent of sample size.
Recently,
this has also been extended for large-scale matrix factorization
\cite{mensch2016dictionary}.  

However,
though CSC is similar to sparse coding, 
the online sparse coding algorithm in \cite{mairal2010online} cannot be directly 
used.
This is because convolution in CSC needs to be performed in the frequency domain for
efficiency.
Moreover, the sizes of history matrices depend on dimensionality of the sparse codes,
which becomes much larger in CSC than in sparse coding.  
Storing the resultant history matrices can be computationally infeasible.

In this paper, we propose a scalable online CSC algorithm for large data sets.  The algorithm,
which will be called Online Convolutional Sparse Coding (OCSC), 
is inspired by the online
sparse coding algorithm of
\cite{mairal2010online}.
It avoids the above-mentioned problems
by reformulating the CSC objective
so that 
convolution 
can be handled easily 
in the frequency domain 
and much smaller history matrices are needed.
We use ADMM to solve the resultant optimization problem. It will be shown that the 
ADMM subproblems have efficient closed-form solutions.
Consequently, to process a given number of samples, 
OCSC has the same time complexity
as state-of-the-art batch CSC methods 
but requires much less space. 
Empirically, as OCSC updates the dictionary after coding each sample, it converges much faster than batch CSC methods.
Theoretical analysis shows that the learned dictionary converges to a stationary point of
the optimization problem.
Extensive experiments show that
convergence of the proposed method is much faster and its reconstruction performance is also better.
Moreover, while existing CSC algorithms can only run on a small number of images,
the proposed method can at least handle ten times more images.

The rest of the paper is organized as follows. Section~\ref{sec:related}
briefly reviews online sparse coding, 
the ADMM,
and batch CSC methods. 
Section~\ref{sec:method} describes the proposed online convolutional sparse coding algorithm.
Experimental results
are presented in Section~\ref{sec:expt}, and
the last section gives some
concluding remarks.

\noindent
{\bf Notations}: 
For vector $a \in\R^m$, its $i$th element is denoted $a(i)$, its $\ell_2$ norm is $\|a\|_2 =
\sqrt{\sum_{i=1}^{m} a^2(i)}$, its $\ell_1$ norm is
$\|a\|_1 = \sum_{i=1}^{m} |a(i)|$, and
$\diag(a)$ reshapes $a$ to a diagonal matrix with elements $a[i]$'s. 
Given another vector $b\in\R^n$, 
the convolution $a*b$ is a vector $c\in\R^{m+n-1}$, with $c(k)=\sum_{j
=\max(1,k+1-n)}^{\min(k,m)} a(j)b(k-j+1)$.
For matrix $A
\in\R^{m\times n}$ with elements
$A(i,j)$'s,
$\text{vec}(A) \in \R^{mn}$ stacks the columns of $A$ to a vector.
Given another matrix $B\in\R^{m\times n}$,
the Hadamard product is
$A\odot B=[A(i,j)B(i,j)]$.
The identity matrix is denoted $I$, and
the conjugate transpose is denoted $(\cdot)^\dagger$.

The Fourier transform that maps from the spatial domain to the frequency
domain
is denoted ${\cal F}$,
and ${\cal F}^{-1}$ is the inverse Fourier transform. For  a
variable $u$ in the spatial domain, its corresponding variable in the frequency domain is
denoted $\tilde{u}$.

\section{Related Works}
\label{sec:related}

\subsection{Online Sparse Coding}
\label{sec:online_sc}

Given $N$ samples $\{x_i,\dots,x_N\}$, where each $x_i\in \mathbb{R}^{P}$, sparse coding
learns an over-complete dictionary $D\in \R^{P \times L} $ of $L$ atoms and sparse codes $\{ z_i\}$ \cite{aharon2006rm}. It can be formulated as the following optimization problem:
\begin{align} 
\label{eq:sc}
\min_{D \in \mathcal{D}, \{z_i\}}
\frac{1}{N}\sum_{i=1}^{N} 
\frac{1}{2}\|x_i-Dz_i\|^2_2 + \beta\|z_i\|_1,
\end{align}
where $\mathcal{D} = \{ D : \NM{ D(:,l) }{2}  \le 1 \text{ for $l=1,\dots,L$}\}$ and $\beta \ge 0$.
Many efficient 
algorithms have been developed for solving 
(\ref{eq:sc}).
Examples include 
K-SVD \cite{aharon2006rm} and active set method \cite{lee2007efficient}. 
However,
they require storing all the samples, which 
can become infeasible
when $N$ is large.

To solve this problem, an online learning algorithm for sparse
coding that processes samples one at a time is proposed in \cite{mairal2010online}. After observing the $t$th sample $x_t$, the sparse code $z_t$ is obtained as
\begin{equation}\label{eq:sc_code}
z_t	=\arg\min_z \frac{1}{2}\| x_t-D_{t-1}z\|^2_2+\beta\|z\|_1, 
\end{equation}
where $D_{t-1}$ is the dictionary obtained at the $(t-1)$th iteration. After obtaining
$z_t$,
$D_t$ is 
updated as
\begin{eqnarray} 
\label{eq:sc_sur}
D_t & = & \arg\min_{D \in \mathcal{D}} \frac{1}{t}\sum_{i=1}^{t}
\frac{1}{2}\| x_i-Dz_i\|^2_2+\beta\|z_i\|_1 
\\	
&=&\arg\min_{D \in \mathcal{D}} \tr(D^\top DA_t^{\text{(osc)}}-2D^\top B_t^{\text{(osc)}}), \label{eq:osc_dl}
\end{eqnarray}
where
\begin{eqnarray}
\label{eq:AB_defi_sc}
A_t^{\text{(osc)}}
& = &\frac{1}{t}\sum_{i=1}^{t}z_iz_i^{\top}\in\mathbb{R}^{L\times L},
\\
\label{eq:AB_defi_sc2}
B_t^{\text{(osc)}}
& =&\frac{1}{t}\sum_{i=1}^{t}x_iz_i^{\top}\in\mathbb{R}^{P\times L}.
\end{eqnarray}
Each column $D_t(:,l)$ in (\ref{eq:osc_dl}) can be obtained by coordinate descent.
$A_t^{\text{(osc)}}$ and $B_t^{\text{(osc)}}$ can also be updated incrementally 
as
\begin{eqnarray}\notag
A_t^{\text{(osc)}} &=& \frac{t - 1}{t} A_{t - 1}^{\text{(osc})} + \frac{1}{t} z_i z_i^{\top},\\
B_t^{\text{(osc)}} &=& \frac{t - 1}{t} B_{t - 1}^{\text{(osc)}} + \frac{1}{t} x_i z_i^{\top}.
\label{eq:upd_hist}
\end{eqnarray} 
Using $A_t^{\text{(osc)}}$ and $B_t^{\text{(osc)}}$, one 
does not need to store
all the samples and codes to update $D_t$.
The whole algorithm is shown in Algorithm~\ref{alg:osc}.

\begin{algorithm}[ht]
	\caption{Online sparse coding \cite{mairal2010online}.}
	\begin{algorithmic}[1]
		\REQUIRE samples $\{x_i\}$.
		\STATE {\bf Initialize}: dictionary $D_0$ as a Gaussian random matrix,
		$A_0^{\text{(osc)}}=\mathbf{0}$, $B_0^{\text{(osc)}}=\mathbf{0}$;
		\FOR{$t = 1,2,\dots, T$}
		\STATE draw $x_t$ from $\{x_i\}$;
		\STATE obtain sparse code $z_t$ using \eqref{eq:sc_code}; 
		\STATE update history matrices $A_t^{\text{(osc)}}, B_t^{\text{(osc)}}$ using
		(\ref{eq:upd_hist});
		\STATE update dictionary $D_t$ using \eqref{eq:osc_dl}
		by coordinate descent;
		\ENDFOR
		\RETURN $D_T$.
	\end{algorithmic}
	\label{alg:osc}
\end{algorithm}

The following assumptions are made in \cite{mairal2010online}.

\begin{assumption} 
	\label{ass:conv}
	\begin{enumerate}
		\item[(A)] Samples $\{x_i\}$ are generated i.i.d. from some distribution with $\|x_i\|_2$ bounded. 
		\item[(B)] The code $z_t$ is unique w.r.t. data $x_t$.
		\item[(C)] The objective
		in \eqref{eq:osc_dl} is strictly convex with lower-bounded Hessians.
	\end{enumerate}
\end{assumption}

\begin{theorem}[\cite{mairal2010online}] \label{th:osc}	
	With Assumption~\ref{ass:conv}, the distance between $D_t$ and the set of stationary points
	of the dictionary learning problem converges almost surely to 0 when $t \rightarrow \infty$.
\end{theorem}

\subsection{Alternating Direction Method of Multipliers (ADMM)}

ADMM \cite{boyd2011distributed} has been popularly used for solving optimization problems of the form
\begin{equation} \label{eq:admm}
\min_{x,y} \; f(x)+g(y) \; : \; Ax+By= c, 
\end{equation} 
where $f, g$ are convex functions, and $A,B$ (resp. $c$) are
constant matrices (resp. vector). 
It first constructs 
the augmented Lagrangian 
of problem \eqref{eq:admm} 
\begin{equation}\label{eq:aug_al}
f(x)+g(y)+\nu^\top(Ax+By-c) +\frac{\rho}{2}\|Ax+By-c\|^2,
\end{equation}
where $\nu$ is the dual variable, and $\rho>0$ is a penalty parameter. 

At the $\tau$th iteration, the values of
$x$ and $y$ (denoted as $x_\tau$ and $y_\tau$)
are updated
by minimizing \eqref{eq:aug_al} w.r.t. $x$ and $y$
in an alternating manner.
Define the scaled dual variable 
$u_\tau={\nu_\tau}/\rho$. The ADMM updates can be written as
\begin{align}
\notag
x_\tau &= \arg \min_{x} f(x)+\frac{\rho}{2}\|Ax+By_{\tau-1}-c+u_{\tau-1}\|_2^2,\\\label{eq:admm_aux}
y_\tau &= \arg \min_{y} g(y)+\frac{\rho}{2}\|Ax_\tau+By-c+u_{\tau-1}\|_2^2,\\\label{eq:admm_dual}
u_\tau &= u_{\tau-1} + Ax_\tau + By_\tau - c.
\end{align}
The above procedure converges to the optimal solution at a rate of $O(1/T)$ \cite{he20121}, 
where $T$ is the number of iterations.

\subsection{Convolutional Sparse Coding}
\label{sec:csc}
Convolutional sparse coding (CSC) learns a dictionary $D \in \R^{M \times K}$ composed of
$K$ filters, each of length $M$,
that can capture the same local pattern at different translated positions of the
samples. This is achieved by replacing the multiplication between dictionary and code
by convolution. While each $x_i$ in sparse coding is represented by a single
code $z_i\in\R^K$, each $x_i$ in CSC is represented by 
$K$ codes stored together in the matrix
$Z_i\in
\R^{P \times K}$.

The dictionary and codes are obtained by solving the following optimization problem:
\begin{equation}\label{eq:csc}
\min_{D \in \mathcal{D}, \{Z_i\}}
\!
\frac{1}{N}
\!\sum_{i=1}^{N}\!
\frac{1}{2}
\| x_i
\!-\!
\sum_{k=1}^{K}D(:,k)
\!*\!
Z_i(:,k) \|^2_2
+ 
\beta\|Z_i\|_1,\!
\end{equation}
where $*$ denotes convolution in the spatial domain.

Convolution can be accelerated in the frequency
domain via the convolution theorem \cite{mallat1999wavelet}:
$\mathcal{F}({D}(:,k) *{Z}(:,k) )=\mathcal{F}({D}(:,k) )\odot\mathcal{F}({Z}(:,k) )$, where ${D}(:,k)$ is first zero-padded to $P$-dimensional.
Hence, 
recent CSC methods \cite{bristow2013fast,heide2015fast,wohlberg2016efficient,sorel2016fast} choose to operate in the frequency domain. 
Let $\tilde{x}_i \equiv \FFT {x_i}$, $\tilde{D}(:,k) \equiv \FFT{D(:,k)}$ and $\tilde{Z}_i(:,k) \equiv \FFT{Z_i(:,k)}$.
\eqref{eq:csc} is reformulated as
\begin{eqnarray}\notag
&\min_{\tilde{D}, \{\tilde{Z}_i\}}&
\frac{1}{N}\sum_{i=1}^{N}
\frac{1}{2P} \| \tilde{x}_i
-
\sum_{k=1}^{K}\tilde{D}(:,k)
\odot
\tilde{Z}_i(:,k) \|^2_2
+\beta\sum_{k=1}^{K}\|\iFFT{\tilde{Z}_i(:,k)}\|_1\!\!
\label{eq:fcsc}
\\
&\text{s.t.}&
\| \HH{\iFFT{\tilde{D}(:,k)}} \|_2^2 \le 1, k =1,\dots,K, \nonumber
\end{eqnarray}
where 
the factor $\frac{1}{P}$ in the objective comes from the Parseval's theorem \cite{chew1995waves},
and
$\mathcal{H}$ is the linear operation that removes the extra $P-M$ dimensions in 
$\iFFT{\tilde{D}(:,k)}$.

Problem~\eqref{eq:fcsc} can be solved by block coordinate descent
\cite{bristow2013fast,heide2015fast,wohlberg2016efficient,sorel2016fast}, 
which 
updates $\{\tilde{Z}_i\}$ and $\tilde{D}$
alternately.

\subsubsection{Updating $\{\tilde{Z}_i\}$}
\label{sec:batch_csc_code}

Given $\tilde{D}$,
the $\{\tilde{Z}_i\}$
can be obtained one by one 
for each $i$th sample
as
\begin{align}
\min_{\tilde{Z}_i,{U}_i}
&~~
\frac{1}{2P} \| \tilde{x}_i
\!-\!
\sum_{k=1}^{K}\tilde{D}(:,k)
\odot
\tilde{Z}_i(:,k) \|^2_2
+\beta\|U_i\|_1
\label{eq:fcsc_code} \\
\text{s.t.} &~~ U_i(:,k)= \iFFT{\tilde{Z}_i(:,k)} , 
\; k =1,\dots,K,
\notag
\end{align} 
where 
$U_i$ is
introduced to decouple the loss and the $\ell_1$-regularizer in \eqref{eq:fcsc}.
This can then be solved by ADMM
\cite{bristow2013fast,heide2015fast,wohlberg2016efficient,sorel2016fast}.

\subsubsection{Updating $\tilde{D}$}
\label{sec:upd_D}

Given $\{\tilde{Z}_i\}$, $\tilde{D}$ can be obtained as
\begin{eqnarray}
\min_{\tilde{D},V}
&
\frac{1}{2NP} \sum_{i=1}^{N} \| \tilde{x}_i
-
\sum_{k=1}^{K}\tilde{D}(:,k)
\odot
\tilde{Z}_i(:,k) \|^2_2
\label{eq:fcsc_dic}
\\\notag
\text{s.t.}
&
\begin{cases}
\FFT{V(:,k)} = \tilde{D}(:,k), 
& k =1,\dots,K,  
\\
\| \HH{(V(:,k))} \|_2^2 \le 1, 
& k =1,\dots,K, 
\end{cases}
\nonumber
\end{eqnarray}
where 
$V$ is
introduced to decouple the loss and constraint
in \eqref{eq:fcsc}.
This can again be solved by using ADMM
\cite{bristow2013fast,heide2015fast,wohlberg2016efficient,sorel2016fast}.

After obtaining $\{\tilde{Z}_i\}$ and $\tilde{D}$, 
the sparse codes can be recovered as $Z_i(:,k) = \iFFT{\tilde{Z}_i(:,k)}$ for $i=1,\dots,N$, and
the dictionary filters as $D(:,k) = \HH{\iFFT{\tilde{D}(:,k)}}$.

The above algorithms 
all need $O(N P K)$ in space.
They differ mainly in how to compute the linear system involved with $\tilde{D}$ in the ADMM subproblems.
FCSC \cite{bristow2013fast} directly solves the subproblem, which takes $O(NK^3P)$ time. 
CBPDN \cite{wohlberg2016efficient} exploits a special structure in the dictionary and
reduces the time complexity to $O(N^2KP)$, which is efficient for small $N$.
The CONSENSUS algorithm \cite{sorel2016fast} utilizes the matrix inverse lemma to reduce the time complexity
to $O(NKP^2)$.
The state-of-the-art is FFCSC \cite{heide2015fast}, which incorporates various linear
algebra techniques (such as Cholesky factorization \cite{jennings1992matrix} and cached factorization \cite{jennings1992matrix}) to reduce the time complexity to $O(NK^2P)$.

\section{Online Convolutional Sparse Coding}
\label{sec:method}

Existing CSC algorithms operate in the batch mode, 
and need to store all the samples and codes which cost $O(N P K)$ space.
This becomes infeasible when the data set is large.
In this section, we will scale up CSC by using online learning.\footnote{After the initial arXiv posting of our paper \cite{wang2017online}, we became aware of some
	very recent independent works that also consider CSC in the online setting
	\cite{degraux2017online,liu2017online}.
	These will be discussed in
	Section~\ref{sec:discuss}.}

After observing the $t$th sample $x_t$, online CSC considers the following optimization
problem which is analogous to \eqref{eq:csc}:
\begin{equation} \label{eq:ocsc0}
\min_{D\in\mathcal{D}, \{Z_i\}}
\frac{1}{t}\sum_{i=1}^{t}
\frac{1}{2}
\| x_i
\!-\!
\sum_{k=1}^{K}D(:,k)*Z_i(:,k) \|^2_2
+ \beta\|Z_i\|_1.
\end{equation} 
To solve problem
(\ref{eq:ocsc0}),
some naive approaches are first considered in Section~\ref{sec:naive}.
The proposed online convolutional sparse coding algorithm is then presented in
Section~\ref{sec:proposed}. It takes the same time  complexity
for one data pass
as state-of-the-art batch CSC algorithms,
but has a much lower space complexity
(Section~\ref{sec:complexity}). 
The convergence properties of the proposed algorithm is discussed in Section~\ref{sec:conv}. 

\subsection{Naive Approaches}
\label{sec:naive}

As in batch CSC, problem~(\ref{eq:ocsc0}) can be solved by alternating minimization w.r.t. the codes and dictionary 
(as in
Section~\ref{sec:csc}). Given the dictionary, the codes are updated as in
(\ref{eq:fcsc_code}). Given the codes, the dictionary is updated by solving the following
optimization subproblem analogous to (\ref{eq:fcsc_dic}):
\begin{eqnarray}
\min_{\tilde{D},V}
&
\frac{1}{2tP} \sum_{i=1}^{t} \| \tilde{x}_i
-
\sum_{k=1}^{K}\tilde{D}(:,k)
\odot
\tilde{Z}_i(:,k) \|^2_2
\label{eq:ocsc}
\\\notag
\text{s.t.}\;\;
&
\begin{cases}
\FFT{V(:,k)} = \tilde{D}(:,k), 
& k =1,\dots,K, 
\\
\| \HH{V(:,k)}\|_2^2 \le 1,
& k =1,\dots,K.
\end{cases}
\notag
\end{eqnarray}
However, solving 
(\ref{eq:ocsc})
as in Section~\ref{sec:upd_D} requires keeping all the samples and
codes, and is computationally expensive on large data sets. 

Alternatively,	the objective in (\ref{eq:ocsc}) can be rewritten as
\begin{equation} \label{eq:tmp1}
\frac{1}{2tP}\sum_{i=1}^{t}\left\|\tilde{x}_i
\! - \!
\dot{D}\dot{z}_i\right\|^2_2, 
\end{equation} 
where 
\begin{equation} \label{eq:dbar}
\dot{D}
=[\text{Diag}(\tilde{D}(:,1)),\dots,\text{Diag}(\tilde{D}(:,K))], 
\end{equation} 
and
$\dot{z}_i=\text{vec}(\tilde{Z}_i)$. 
This is of the
form in \eqref{eq:sc_sur}. Hence, we may attempt to reuse the online sparse coding in
Algorithm~\ref{alg:osc}, and thus avoid storing all the samples and codes. However, recall
that online learning the dictionary is possible because
one can summarize
the observed samples 
into the history matrices
$A_t^{\text{(osc)}},B_t^{\text{(osc)}}$ in \eqref{eq:AB_defi_sc}, \eqref{eq:AB_defi_sc2}. For (\ref{eq:tmp1}), the history matrices become
\begin{eqnarray} \label{eq:hisold}
{A}_t^{\text{(naive)}} &=& \frac{1}{t}\sum_{i=1}^{t}\dot{z}_i\dot{z}_i^{\dagger}\in\mathbb{R}^{PK\times PK},\\
{B}_t^{\text{(naive)}} &=&\frac{1}{t}
\sum_{i=1}^{t}\tilde{x}_i\dot{z}_i^{\dagger}\in\mathbb{R}^{P\times PK}.\nonumber
\end{eqnarray}
In typical CSC
applications, the number of image pixels $P$ is at least in the tens of thousands, and $N$
may only be in the thousands. 
Hence, the
$O(K^2 P^2)$ space required for storing ${A}_t^{\text{(naive)}}$ and ${B}_t^{\text{(naive)}}$ is even
higher than the $O(NKP)$ space required for batch methods.

\subsection{Proposed Algorithm}\label{sec:proposed}
Note that
$\dot{D}$ in (\ref{eq:dbar}) is composed of a number of diagonal matrices.
By utilizing this special structure,
the following Proposition 
rewrites the objective in (\ref{eq:ocsc}) 
so that much smaller history matrices can be used.

\begin{proposition} \label{pr:reorder}
	The objective in (\ref{eq:ocsc}) is equivalent to the following apart from a constant:
	\begin{eqnarray} \label{eq:sur_prop}
	\frac{1}{2P}\sum_{p=1}^{P}
	\tilde{D}(p,:)\left(\frac{1}{t}\sum_{i=1}^{t}\tilde{Z}_i^{\dagger}(:,p)\tilde{Z}_i(p,:)
	\right) \tilde{D}^{\dagger}(:,p)
	-\frac{1}{P}\sum_{p=1}^{P}\tilde{D}(p,:)\left(\frac{1}{t}\sum_{i=1}^{t}\tilde{x}_i(p)\tilde{Z}_{i}^{\dagger}(:,p)\right),
	\end{eqnarray}
\end{proposition}
The proof is in Appendix~\ref{app:reorder}. Obviously, each $\tilde{D}(p,:)$ in
(\ref{eq:sur_prop}) can then be independently optimized.
This avoids directly handling the
much larger $P\times PK$ matrix $\dot{D}$ in 
(\ref{eq:dbar}).
Let
\begin{eqnarray}\notag
{A}^p_t&=&\frac{1}{t}\sum_{i=1}^{t}\tilde{Z}_i^{\dagger}(:,p)\tilde{Z}_i(p,:)\in\mathbb{R}^{K\times K},\\
\notag
{b}^p_t&=&\frac{1}{t}\sum_{i=1}^{t}\tilde{x}_i(p)\tilde{Z}_{i}^{\dagger}(:，p)\in\mathbb{R}^{K}
\end{eqnarray}
in 
(\ref{eq:sur_prop}).
The total space required 
for $\{{A}^p_t,{b}^p_t \}$
is $O(K^2P)$, which is much smaller than the $O(K^2P^2)$ space for storing
${A}_t^{\text{(naive)}}$ 
in (\ref{eq:hisold}). Moreover,
as in (\ref{eq:upd_hist}), 
${A}^p_t$ and ${b}^p_t$
can be updated incrementally as
\begin{align}\label{eq:A_update}
{A}^p_t&=\left(1-\frac{1}{t} \right){A}^p_{t-1}+\frac{1}{t}\tilde{Z}_t^{\dagger}(:,p)\tilde{Z}_t(p,:),\\
{b}^p_t&=
\left(1-\frac{1}{t} \right){b}^p_{t-1}+\frac{1}{t}\tilde{x}_t(p)\tilde{Z}_{t}^{\dagger}(:，p).
\label{eq:B_update}
\end{align}
The dictionary and codes 
can then be efficiently updated in an alternating manner
as follows.

\subsubsection{Updating the Dictionary}
\label{sec:updateDict}

With the codes fixed,
using Proposition~\ref{pr:reorder},
the dictionary can be updated by solving the following optimization problem:
\begin{align}
(\tilde{D}_t,V_t) 
& =
\arg\min_{\tilde{D},V}
\frac{1}{2P}\sum_{p=1}^{P}\!\!
\tilde{D}(p,:){A}^p_t \tilde{D}^{\dagger}(:,p) \!-\!2\tilde{D}(p,:){b}^p_t
\label{eq:ocsc_dic} \\
&	\text{s.t.} \; \FFT{V(:,k)} = \tilde{D}(:,k), k =1,\dots,K, 
\quad
\| \HH{V(:,k)} \|_2^2 \le 1, k =1,\dots,K. \nonumber
\end{align}
This can solved using ADMM.  At the $\tau$th ADMM iteration, let $\Theta_{t,\tau}$ be
the ADMM dual variable.  The following shows the update equations for $\tilde{D}$ and
$V$.


\noindent
{\bf Updating $\tilde{D}_{t,\tau}$:}
From \eqref{eq:sur_prop}, 
$\tilde{D}_{t,\tau}$ can be updated by solving the following subproblem:
\begin{align*}
\tilde{D}_{t,\tau}
=  \arg\min_{\tilde{D}}
\frac{1}{2P}\sum_{p=1}^{P}
\tilde{D}(p,:){A}^p_{t} \tilde{D}^{\dagger}(:,p)-2\tilde{D}(p,:){b}^p_t
+\frac{\rho}{2}
\sum_{k=1}^{K}\|\tilde{D}(:,k)-\tilde{V}_{t,\tau-1}(:,k)+\Theta_{t,\tau-1}(:,k)\|_2^2,
\end{align*}
where
$\tilde{V}_{t,\tau-1}(:,k)\equiv \FFT {{V}_{t,\tau-1}(:,k)}$.
Note that
$\|\tilde{D}\|_F^2=\sum_{p=1}^{P}\|\tilde{D}(p,:)\|_2^2=\sum_{k=1}^{K}\|\tilde{D}(:,k)\|_2^2$.
Hence,
$
\sum_{k=1}^{K}\|\tilde{D}(:,k)-\tilde{V}_{t,\tau-1}(:,k)+\Theta_{t,\tau-1}(:,k)\|_2^2=
\sum_{p=1}^{P}\|\tilde{D}(p,:)-\tilde{V}_{t,\tau-1}(p,:)+\Theta_{t,\tau-1}(p,:)\|_2^2$.
The rows of $\tilde{D}_{t,\tau}$ 
can then be obtained separately as
\begin{eqnarray}
\tilde{D}_{t,\tau}(p,:) \!\!\!\! &=& 
\!\!\!\! \min_{\tilde{D}(p,:)}		
\frac{1}{2P}
(\tilde{D}(p,:){A}^p_t \tilde{D}^{\dagger}(:,p) -2\tilde{D}(p,:){b}^p_t) 
+\frac{\rho}{2}
\|\tilde{D}(p,:)-\tilde{V}_{t,\tau-1}(p,:)+\Theta_{t,\tau-1}(p,:)\|_2^2\nonumber\\
&=& \!\!\!\! 
({b}^{p\dagger}_t \!+\! \rho P\tilde{V}_{t,\tau-1}(p,:) \!-\!\rho P\Theta_{t,\tau-1}(p,:)){C}^p_t,
\label{eq:sub_d_sol}
\end{eqnarray}
where ${C}^p_t = ({A}^p_t+\rho P I)^{-1}$.
With $\{C^p_t\}$, we do not need to store $\{A^p_t\}$.

Computing the matrix inverse 
${C}^p_t \in\C^{K\times K}$
takes $O(K^3)$ time. 
This can be simplified by noting from (\ref{eq:A_update})
that ${A}^p_t+\rho P I$ is the sum of rank-1 matrices
and a (scaled) identity matrix. Using the Sherman-Morrison formula\footnote{Given an invertible square matrix 
	$A$ 
	and 
	vectors
	$u,v$,
	$\left( A+ u v^{\top} \right)^{-1}=A^{-1}- \left( 1+v^{\top} A^{-1} u \right)^{-1}\left( A^{-1} u v^{\top} A^{-1} \right)$.}
\cite{jennings1992matrix}, we have
\begin{equation} \label{eq:C_update}
{C}^p_t= 
\left\{
\begin{array}{ll}
\frac{1}{\rho P}I-\frac{1}{\rho P}\frac{\tilde{Z}_t^{\dagger}(:,p)\tilde{Z}_t(p,:)}{\rho
P+\tilde{Z}_t(p,:)\tilde{Z}_t^{\dagger}(:,p)}
&
t=1 \\
\frac{t}{t-1}
\left[{C}^p_{t-1}-\frac{{C}^p_{t-1}\tilde{Z}_t^{\dagger}(:,p)\tilde{Z}_t(p,:){C}^p_{t-1}}{(t-1)+\tilde{Z}_t(p,:){C}^p_{t-1}\tilde{Z}_t^{\dagger}(:,p)}
\right]
& t>1
\end{array}
\right..
\end{equation}
This takes $O(K^2)$, instead of $O(K^3)$, time.


\noindent
{\bf Updating $V_{t,\tau}$:}
From \eqref{eq:admm_aux}, each column $V_{t,\tau}(:,k)$ can be updated as
\begin{align*}
\min_{V(:,k)}
\frac{\rho}{2}\|\tilde{D}_{t,\tau}(:,k) \! - \! \FFT {V(:,k)} \! + \! \Theta_{t,\tau-1}(:,k)\|_2^2,
\quad
\text{s.t.} \| \HH{V(:,k)} \|_2^2\le 1. 
\end{align*}
It has the following closed-form solution.

\begin{proposition}[\cite{parikh2014proximal}] \label{pr:sub_v_sol}
$V_{t,\tau}(:,k) = \alpha/\max( \|\alpha\|_2, 1)$,
where $\alpha = \HH{ \iFFT{\tilde{D}_{t,\tau}(:,k)+\Theta_{t,\tau-1}(:,k)} }$. 
\end{proposition}
Finally, the dual variables $\Theta_{t,\tau}$
is updated as in \eqref{eq:admm_dual}.
The whole dictionary update procedure ($\text{DictOCSC}$)
is shown in Algorithm~\ref{alg:ocsc_dl}.
As \eqref{eq:ocsc_dic} is convex,
convergence
to the globally optimal solution is guaranteed \cite{he20121}.

\begin{algorithm}[ht]
\caption{DictOCSC($\tilde{D}_{t-1},\{b^p_t\},\{C^p_t\}$).}
\begin{algorithmic}[1]
\REQUIRE initial dictionary $\tilde{D}_{t-1}$, $\{{b}^p_t\}$, $\{{C}^p_t\}$;
\STATE {\bf Initialize}: $\tilde{D}_{t,0}=\tilde{D}_{t-1}$, $V_{t,0}=\mathbf{0}$, $\Theta_{t,0}=\mathbf{0}$;
		\FOR{$\tau = 1,2,\dots, J$}	
\STATE update $\{\tilde{D}_{t,\tau}(1,:),\dots, \tilde{D}_{t,\tau}(P,:)\}$ using (\ref{eq:sub_d_sol});
\STATE update $\{V_{t,\tau}(:,1), \dots, V_{t,\tau}(:,K)\}$ using Proposition~\ref{pr:sub_v_sol};
\STATE update $\{\Theta_{t,\tau}(:,1), \dots, \Theta_{t,\tau}(:,K)\}$ as
$\Theta_{t,\tau}(:,k) = \Theta_{t,\tau-1}(:,k) + \tilde{D}_{t,\tau}(:,k) -\FFT
{V_{t,\tau}(:,k)}$;
		\ENDFOR
		\RETURN $\tilde{D}_{t,J}$.
	\end{algorithmic}
	\label{alg:ocsc_dl}
\end{algorithm}

	In batch CSC methods, its dictionary update in \eqref{eq:fcsc_dic} is also based on ADMM
	(Section~\ref{sec:upd_D}). However, our dictionary update step first reformulates the
	objective as in \eqref{eq:sur_prop}. This enables each ADMM subproblem to be solved with a
	much lower space complexity ($O(K^2P)$ vs $O(NKP)$ for the state-of-the-art \cite{heide2015fast}) but
	still with the same iteration time complexity 
	(i.e., $O(NK^2P+NKP\log P)$).

\subsubsection{Updating the Code}\label{sec:ocsc_code}

Given the dictionary, 
as the codes for different samples are independent,
they can be updated one by one as in batch CSC methods (Section~\ref{sec:batch_csc_code}).

The whole algorithm, which will be called
Online Convolutional Sparse Coding (OCSC),
is shown in Algorithm~\ref{alg:ocsc}.

\begin{algorithm}[H]
	\caption{Online convolutional sparse coding (OCSC).}
	\begin{algorithmic}[1]
		\REQUIRE samples $\{x_i\}$.
		\STATE {\bf Initialize}: 
		dictionary $\tilde{D}_0$ as a Gaussian random matrix,
		$\{ {C}_0^p \} = \mathbf{0}$,
		$\{ {b}_0^p \} = \mathbf{0}$;
		\FOR{$t = 1, 2,\dots, T$}
		\STATE $\tilde{x}_t = \FFT {x_t}$, where $x_t$ is drawn from $\{x_i\}$;
		\STATE obtain $\tilde{Z}_t$ using \eqref{eq:fcsc_code};
		\STATE update $\{ {b}^1_t,\dots,
		{b}^P_t
		\}$ using \eqref{eq:B_update};
		\STATE update $\{ {C}^1_t,\dots,
		{C}^P_t \}$ using \eqref{eq:C_update};
		\STATE $\tilde{D}_t=\text{DictOCSC}(\tilde{D}_{t-1},\{b^p_t\},\{C^p_t\})$.
		\ENDFOR
		\FOR{$k = 1, 2,\dots, K$}
		\STATE
		$D_T(:,k) = \HH{ \iFFT{\tilde{D}_T(:,k)} }$;
		\ENDFOR
		\RETURN $D_T$.
	\end{algorithmic}
	\label{alg:ocsc}
\end{algorithm}

\subsection{Complexity Analysis}
\label{sec:complexity}	

In Algorithm~\ref{alg:ocsc}, the space requirement is dominated by $ \{ C^p_t \} $, which
takes $O(K^2P)$ space. For one data pass which precesses $N$ samples,
updating $ \{ C^p_t \} $ and $ \{ b^p_t \} $ 
takes $O(NK^2P)$ time, dictionary update takes $O(NK^2P)$ time, 
code update takes $O(NKP)$ time,
and
FFT/inverse FFT
takes $O(NKP\log P)$ time.
Hence, one data pass takes a total of $O(NK^2P +NKP\log P)$ time.

A comparison 
with the existing batch CSC algorithms is shown in 
Table~\ref{tab:cost}.
As can be seen, the proposed algorithm takes the same time complexity for one data pass as the state-of-the-art FFCSC algorithm,
but has a much lower space complexity
($O(K^2P)$ instead of $O(NKP)$). 
\begin{table}[htbp]
	\caption{Comparing the proposed online CSC algorithm with existing batch CSC algorithms.
	}
	\centering
	\begin{tabular}{c|c|c|c|c} \hline
		&
		batch/online & convolution operation & space & time for one data pass \\\hline
		DeconvNet \cite{zeiler2010deconvolutional}&batch &spatial&$O(NKP)$ &$O(NK^2P^2M)$ \\\hline
		FCSC \cite{bristow2013fast}&batch &frequency&$O(NKP)$ &$O(NK^3P+NKP\log P)$ \\\hline
		FFCSC \cite{heide2015fast}&batch&frequency&$O(NKP)$ &$O(NK^2P+NKP\log P)$  \\\hline
		CBPDN \cite{wohlberg2016efficient}&batch &frequency&$O(NKP)$ & $O(N^2KP+NKP\log P)$ \\\hline
		CONSENSUS \cite{sorel2016fast}&batch&frequency&$O(NKP)$ &$O(NKP^2+NKP\log P)$  \\\hline	
		OCSC &	online&frequency&$O(K^2P)$ &$O(NK^2P+NKP\log P)$ \\\hline	
	\end{tabular}
	\label{tab:cost}
\end{table}

\subsection{Convergence}
\label{sec:conv}

In this section,
we show that Algorithm~\ref{alg:ocsc} outputs a stationary point of the CSC problem \eqref{eq:ocsc0} when $t \rightarrow \infty$.
This is achieved by connecting Algorithm~\ref{alg:ocsc} to a direct application of Algorithm~\ref{alg:osc} on \eqref{eq:ocsc0}. 

The convolution operation in the spatial domain can be written as matrix multiplication
\cite{bristow2013fast,zeiler2010deconvolutional}. Specifically,
\begin{equation}\label{eq:csc2sc}
D(:,k) * Z(:,k) = \mathcal{T}(D(:,k)) Z(:,k),
\end{equation}
where
$\mathcal{T}(x)$ is a linear operator which maps a vector to
its associated Toeplitz matrix. Specifically,
\begin{eqnarray*}
	\mathcal{T}(D(:,k))
	= \left[
	\begin{array}{ccccc}
		D(1,k)  &0 			    &0 			 			&		    			  &D(2,k) \\
		D(2,k)  &D(1,k)  	 &0		     		     &\ddots  				&D(3,k)\\
		D(3,k)  &D(2,k) 	 &D(1,k)    		  &  		  				&D(4,k)\\
		\vdots 	&\vdots		 &\vdots    		   &\ddots 			&\vdots\\
		D(M,k)  &D(M-1,k) ]&D(M-2,k)           &                       &\\
		0 		  &D(M,k)      &D(M-1,k)           &                       &\\
		\vdots  &\vdots      &\vdots              &\ddots              &\vdots\\
		0         & 0             &0              		&                        &  D(1,k) 
	\end{array}
	\right].
\end{eqnarray*}
The number of columns in
		$\mathcal{T}(D(:,k))$
is equal to the dimension of $Z(:,k)$ (i.e., $P$).

Let $\bar{D}_k = \mathcal{T}(D(:,k))\in\R^{P\times P}$, 
$\mathcal{T}^{-1}$ be the inverse operator of $\mathcal{T}$ which maps $\bar{D}_k$ back to $D(:,k)$, 
$\bar{D} = \left[ \bar{D}_1, \dots, \bar{D}_K \right]$
and $\bar{z}_i = \text{vec}(Z_i)$.
Problem~\eqref{eq:ocsc0} can be rewritten as
\begin{align}
\min_{\bar{D} \in \mathcal{S}, \{\bar{z}_i\}} 
\frac{1}{t}\sum_{i=1}^{t}\frac{1}{2}
\|x_i - \bar{D} \bar{z}_i\|^2_2 
+ \beta\| \bar{z}_i \|_1,
\label{eq:cscequiv} 
\end{align}
where 
\begin{equation} \label{eq:s}
\mathcal{S} = \left\lbrace \bar{D}_k : \| \mathcal{T}^{-1} ( \bar{D}_k ) \|_2^2 \le 1 \right\rbrace.
\end{equation} 
Thus,
the objective \eqref{eq:cscequiv} is of the same form as that in (\ref{eq:sc_sur}).
However, a direct use of Algorithm~\ref{alg:osc} is not feasible. 
First, the feasible region $\mathcal{S}$ in \eqref{eq:s} is more complex,
and coordinate descent cannot be used
as there is no simple projection to $\mathcal{S}$. 
Second, 
as $\bar{z}_i$ is of length $KP$, 
the corresponding history matrices (analogous to those in \eqref{eq:AB_defi_sc}) require
$O(K^2P^2)$ space.


Though a direct application of Algorithm~\ref{alg:osc} is not practical,
Theorem~\ref{th:osc} still holds.
Indeed,
Theorem~\ref{th:osc} can be further extended by relaxing its
feasible region 
$\mathcal{C}$ 
on $D$.
As
discussed in \cite{mairal2010online},
$\mathcal{C}$ can be,
for example, 
$\{ D : \| D(:, l) \|_2 \le 1 \;\text{and}\; D(i, j) \ge 0 \}$.
It is mentioned in 
\cite{mairal2010online}
that $\mathcal{C}$ has to be a union of independent constraints on each column of $D$.
However, 
this only serves to facilitate the use of coordinate descent (step~6 in
Algorithm~\ref{alg:osc}), but  is
not required in the proof.
In general,
Theorem~\ref{th:osc} holds when $\mathcal{C}$ is bounded, convex, and $\mathcal{C} \subseteq \mathcal{D}$. 





The following Lemma shows that $\mathcal{S}$ in \eqref{eq:s} satisfies the
conditions.
The proof is in Appendix~\ref{app:lemD}.
Thus, Theorem~\ref{th:osc} also holds for Algorithm~\ref{alg:ocsc}, and a stationary point
of problem~\eqref{eq:ocsc0} can be obtained.

\begin{lemma}
\label{lem:bndD}
$\mathcal{S}$ 
in \eqref{eq:s} 
is bounded, convex, and a subset of $\mathcal{D}$.
\end{lemma}

%
%

\subsection{Discussion with \cite{degraux2017online,liu2017online}}
\label{sec:discuss}

Here, we discuss the very recent works of \cite{degraux2017online,liu2017online}
which also consider online learning of the dictionary in CSC.
Extra experiments are performed in Section~\ref{sec:extraExp},
which shows our method is much faster than them.

In the online convolutional dictionary learning (OCDL) algorithm \cite{degraux2017online},
convolution is performed in the spatial domain.
They started with the observation that 
convolution
is commutative. Hence, for the summation
	$\sum_{k=1}^{K}D(:,k) * Z_i(:,k)$ in 
\eqref{eq:ocsc0}, we have
\begin{eqnarray}\label{eq:equiv_conv}
	\sum_{k=1}^{K}D(:,k) * Z_i(:,k)
	= \sum_{k=1}^{K} Z_i(:,k) *D(:,k)
	= \sum_{k=1}^{K}\mathcal{T}(Z_i(:,k))D(:,k)=\bar{Z}_i\bar{d},
\end{eqnarray}
where 
$\bar{Z}_i = \left[ \mathcal{T}(Z_i(:,1)),\dots,\mathcal{T}(Z_i(:,K)) \right]$ with $\mathcal{T}(Z_i(:,k))\in\R^{P\times M}$ 
and 
$\bar{d}=\text{vec}(D)$.
\eqref{eq:ocsc0} can then be rewritten as 
\begin{eqnarray}
\label{eq:odl_de}
\min_{\bar{d}, \{\bar{Z}_i\}}
\frac{1}{2t}\sum_{i=1}^{t}\left\|{x}_i
- 
\bar{Z}_i\bar{d}\right\|^2_2
+
\lambda\|\bar{Z}_i\|_1
\quad
\text{s.t.}\quad
\|D(:,k)\|_2^2 \le 1, k =1,\dots,K, 
\end{eqnarray}
where $\lambda = \beta/P$ (as each $Z_i(:,k)$ is repeated $P$ times in the Toeplitz matrix
$\bar{Z}_i$).
Using \eqref{eq:odl_de}, the history matrices are constructed as
\begin{eqnarray*}
	{A}_t^{\text{(ocdl1)}}
	&=&\frac{1}{t}\sum_{i=1}^{t}\bar{Z}_i^{\top}\bar{Z}_i \in\mathbb{R}^{KM\times KM},\\
	{b}_t^{\text{(ocdl1)}} &=&\frac{1}{t}\sum_{i=1}^{t}\bar{Z}_i^{\top}x_i \in\mathbb{R}^{KM}.
\end{eqnarray*}
Recall that $M$ is the length of filter $D(:,k)$ when CSC is solved in the spatial domain.
The space complexity of \cite{degraux2017online} is dominated by $\bar{Z}_i$ (which
takes $O(KPM)$ space) or ${A}_t^{\text{(ocdl1)}}$ (which takes $O(K^2M^2)$ space), depending
on the relative sizes of $P$ and $KM$. 
Though this is comparable to our $O( K^2 P )$ space requirement, its
time complexity  is
much larger.
For one data pass, 
convolution in the spatial domain takes $O(NKPM)$ time and updating the history
matrices above takes $O( NK^2PM^2 )$ time.
The 
dictionary 
update in total
takes $O(NK^2PM^2 + NKPM)$ time. In contrast,
the proposed algorithm takes
$O(NK^2P+NKP\log P)$ time. 
In the experiments, $M = 11\times 11$, and $P$ ranges from $32\times32$ to
$500\times500$.
Thus,
the algorithm in \cite{degraux2017online} is much more expensive.

The algorithm in 
\cite{liu2017online}, also called online convolutional dictionary learning,
considers
the frequency domain
and solves problem~\eqref{eq:ocsc} as in the proposed method.
First, 
they rewrite \eqref{eq:ocsc} as
\begin{eqnarray}\label{eq:ocdl2}
\min_{\dot{d}, \{\dot{Z}_i\}}
\frac{1}{2tP}\sum_{i=1}^{t}\left\|\tilde{x}_i
- 
\dot{Z}_i\dot{d}\right\|^2_2
+
\beta\|\dot{Z}_i\|_1
\quad\text{s.t.}\quad
\| \HH{\iFFT{\tilde{D}(:,k)}} \|_2^2 \le 1, k =1,\dots,K, \nonumber
\end{eqnarray}
where $\dot{Z}_i=[\text{Diag}(\tilde{Z}_i(:,1)),\dots,\text{Diag}(\tilde{Z}_i(:,K))]$, 
and
$\dot{d}=\text{vec}(\tilde{D})$. 
The history matrices 
are then constructed  as
\begin{eqnarray*}
	{A}_t^{\text{(ocdl2)}}
	&=&\frac{1}{t}\sum_{i=1}^{t}\dot{Z}_i^{\top}\dot{Z}_i \in\mathbb{R}^{KP\times KP},
	\\
	{b}_t^{\text{(ocdl2)}} &=&\frac{1}{t}\sum_{i=1}^{t} \dot{Z}_i^{\top}\tilde{x}_i \in\mathbb{R}^{KP}.
\end{eqnarray*}
They proposed to store ${A}_t^{\text{(ocdl2)}}$ and the history matrices 
in sparse format. These take $O(K^2P)$ space,
and is the same as 
our $\{C^p_t\}$ in
\eqref{eq:C_update} which are stored in dense format.
However, though their $O(\cdot)$ are the same, 
storing sparse matrix requires 2-3 times more space than 
storing dense matrix with the same number of nonzero entries,
as
each nonzero entry in a sparse matrix needs to be kept in the
compressed sparse row (CSR)\footnote{For an introduction to the CSR format, interested readers are referred to \url{http://www.netlib.org/utk/people/JackDongarra/etemplates/node373.html}.}
format \cite{davis2006direct}.
Moreover, though
\cite{liu2017online} and the proposed method
take $O( NK^2P )$ time
to update the history matrices,
using sparse matrices as in
\cite{liu2017online} 
is empirically slower
\cite{davis2006direct}.
Preliminary experiments 
show that with $K=100$ filters, the
proposed algorithm is $5$ times faster on an 
$P=100\times100$ 
image 
and $10$ times
faster on an 
$P = 200\times200$
image.

Moreover, we use ADMM for optimization, while 
\cite{degraux2017online,liu2017online} use the FISTA algorithm \cite{beck2009fast}.
Empirically, 
ADMM has been shown to be faster than
FISTA 
on solving the CSC problem
\cite{wohlberg2016efficient}.
We also empirically verify this point in Section~\ref{sec:appadmmfis}.

\section{Experiments}
\label{sec:expt}

In this section, 
experiments are performed on six image data sets
(Table~\ref{tab:data_stat}). The \textit{Fruit}, \textit{City} and \textit{House} data
sets\footnote{\url{http://www.cs.ubc.ca/labs/imager/tr/2015/FastFlexibleCSC/}.} are standard
benchmarks in CSC \cite{bristow2013fast,heide2015fast,zeiler2010deconvolutional}. 
However, they are relatively small.
Hence,
we include three larger data sets: (i)
\textit{Flower},\footnote{\url{http://www.robots.ox.ac.uk/~vgg/data/flowers/102/}.}
which 
contains images from 102 flower categories
\cite{nilsback2008automated}; (ii)
\textit{Dog},\footnote{\url{http://vision.stanford.edu/aditya86/ImageNetDogs/}.}
which contains images of 120 breeds of dogs
\cite{khosla2011novel};
and 
(iii) 
\textit{CIFAR-10},\footnote{\url{https://www.cs.toronto.edu/~kriz/cifar.html}.}
which
contains 
images from 10 general object classes
\cite{krizhevsky2009learning}.
The training images are used to learn the
dictionary, which is then used to 
reconstruct
the test set images.
We use the default training and test splits provided.
The images are preprocessed as in 
\cite{heide2015fast,zeiler2010deconvolutional}. We convert each image to grayscale, and
perform local contrast normalization. Edge-tapering is used to blur the edges of the
samples with a random Gaussian filter.

\begin{table}[ht]
	\caption{Summary of data sets used.}
	\centering
	\begin{tabular}{c|ccc } \hline
		& dimension ($P$) & \#training  images& \#testing images \\\hline
		\it Fruit &100$\times$100 &10  & 4  \\\hline
		\it City & 100$\times$100 & 10  &4   \\\hline
		\it House& 100$\times$100 & 100 & 4  \\\hline
		\it Flower& 500$\times$500 & 2040& 6149  \\\hline
		\it Dog& 224$\times$224 & 12000& 8580  \\\hline	
		\it CIFAR-10& 32$\times$32 & 50000 &10000  \\\hline
	\end{tabular}
	\label{tab:data_stat}
\end{table}	
The proposed OCSC is compared with the following (batch) CSC methods:
\begin{enumerate}
	\item Deconvolutional networks (DeconvNet)
	\cite{zeiler2010deconvolutional};
	\item Fast convolutional sparse coding (FCSC) \cite{bristow2013fast};
	\item Fast and flexible convolutional sparse coding (FFCSC) \cite{heide2015fast};
	\item Convolutional basis pursuit denoising (CBPDN) \cite{wohlberg2016efficient};
	\item The global consensus ADMM (CONSENSUS) algorithm \cite{sorel2016fast}.
\end{enumerate}
All the codes are in Matlab and obtained
from the respective authors (except FCSC).\footnote{DeconvNet is from \url{http://www.matthewzeiler.com/}, FFCSC from \url{http://www.cs.ubc.ca/labs/imager/tr/2015/FastFlexibleCSC/},
	CBPDN from \url{http://brendt.wohlberg.net/software/SPORCO/}, and CONSENSUS is from
	\url{http://zoi.utia.cas.cz/convsparsecoding}. The code of
	FCSC is not 
	available, so we use the code in \cite{sorel2016fast} (\url{http://zoi.utia.cas.cz/convsparsecoding}).} 
We do not compare with 
\cite{degraux2017online,liu2017online}, as
their codes are not publicly available.

We 
follow the hyperparameter setting in \cite{heide2015fast}, and set $\beta=1$, $M=11$ and
$K=100$.
The batch methods are stopped
when the
relative changes in $\{Z_i\}$ (i.e.,
$\frac{1}{N}\sum_{i=1}^N\|Z_i-Z_i^{\text{old}}\|_2/\|Z_i\|_2$) and
$D$ (i.e.,
$\|D-D^{\text{old}}\|_2/\|D\|_2$) are both smaller than ${10}^{-3}$.
As for the proposed online method, 
it is stopped when the
relative changes in $Z_t$ (i.e.,
$\|Z_t-Z_t^{\text{old}}\|_2/\|Z_t\|_2$) and
$D$ (i.e.,
$\|D-D^{\text{old}}\|_2/\|D\|_2$) are both smaller than ${10}^{-3}$.
Experiments are run on a PC with Intel i7 4GHz CPU with 32GB memory.

To compare the empirical convergence of various methods, we monitor the 
test set 
objective 
as in \cite{mairal2010online,mensch2016dictionary}. 
This is obtained as
\begin{equation*}
\min_{\{Z_i\}}
\frac{1}{|\Omega|}\sum_{x_i \in \Omega }
\frac{1}{2}
\| x_i-\sum_{k=1}^{K}D_t(:,k)*Z_i(:,k) \|^2_2
+ \beta\|Z_i\|_1,
\end{equation*}
where $\Omega$ is the test set, 
and $D_t$ is the dictionary learned after the $t$th 
data pass.
As for image reconstruction quality, we use the peak signal-to-noise ratio on the test set \cite{heide2015fast}: 
\begin{equation*}
\text{PSNR}=\frac{1}{|\Omega|}\sum_{x_i \in \Omega }10\log_{10}\left(
\frac{255^2P}{\|{x}^{\text{rec}}_i-{x}_i\|_2^2 } \right) ,
\end{equation*}
where
${x}^{\text{rec}}_i= \sum_{k=1}^{K}D_T(:,k)*Z_i(:,k)$
is the reconstructed image for $x_i$ by using the final dictionary $D_T$.
To reduce statistical variability, we repeat the experiment five times by using different
dictionary initializations and orders to present images to the algorithm.
The PSNR is then averaged over these five 
repetitions.

\subsection{Small Data Sets}
\label{sec:expts_small}
\begin{figure}[ht]
	\centering
	\subfigure[\textit{Fruit}. \label{fig:test_obj_fruit}]{\includegraphics[width=0.32\textwidth]{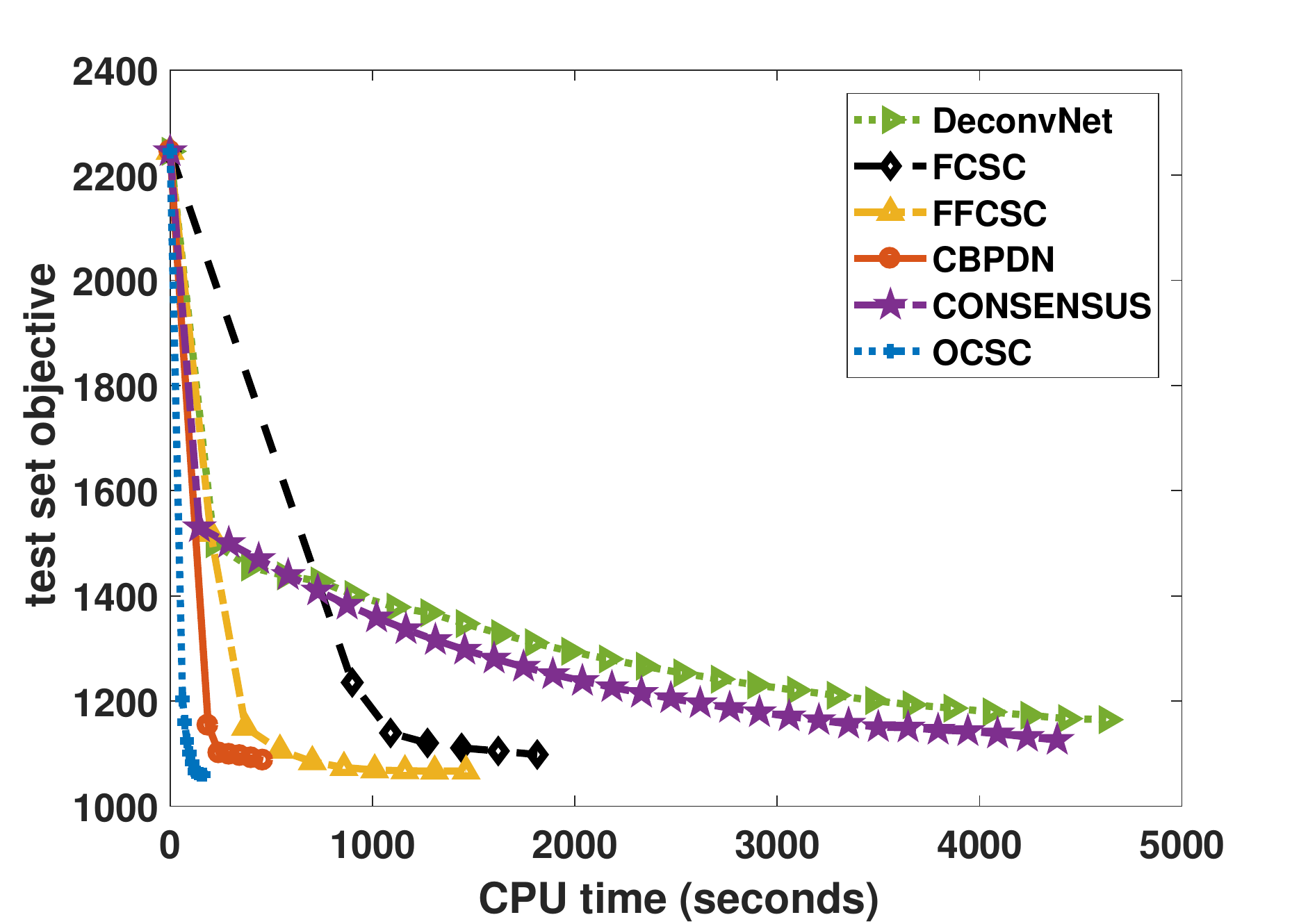}}
	\subfigure[\textit{City}. \label{fig:test_obj_city}]{\includegraphics[width=0.32\textwidth]{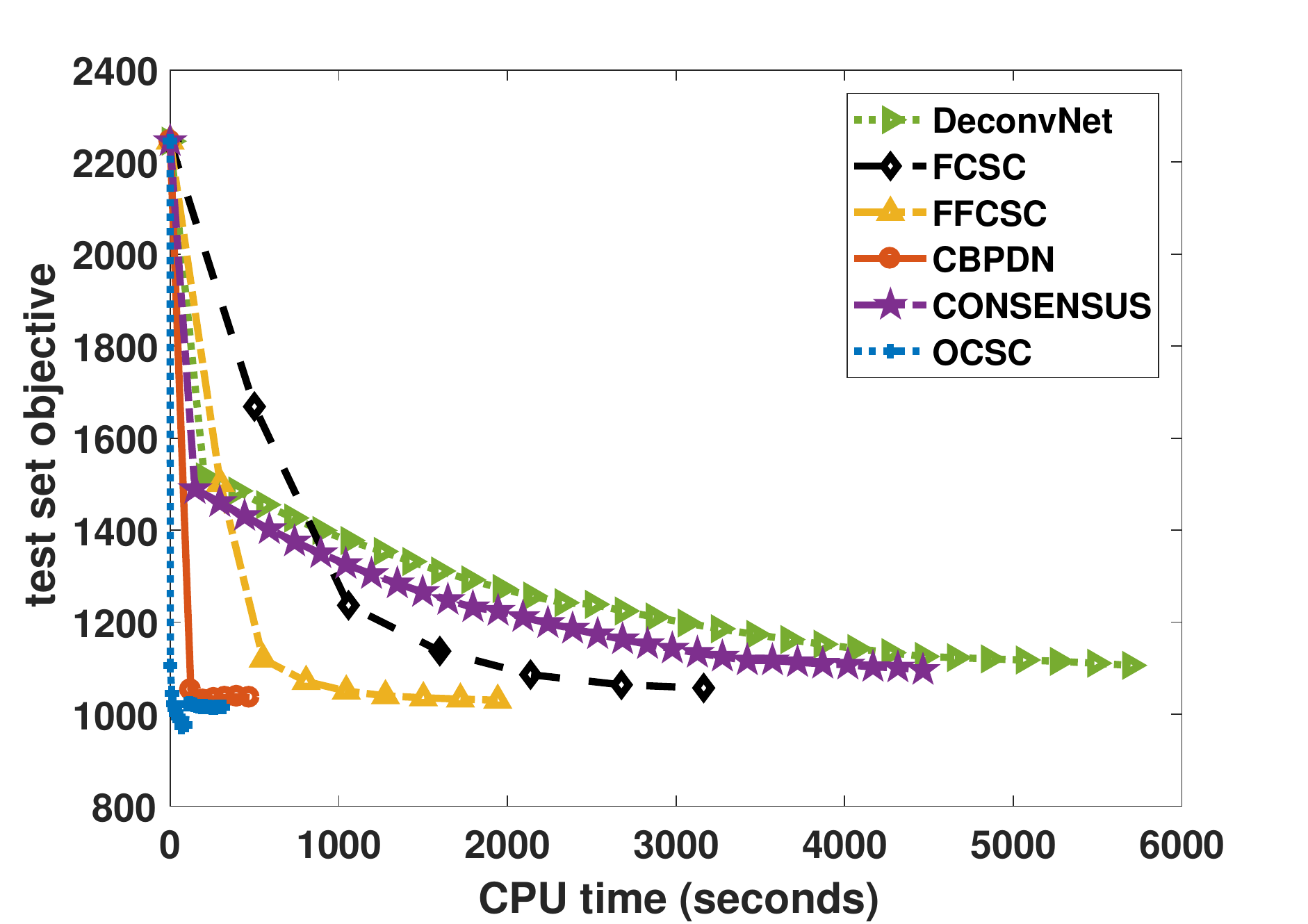}}
	\subfigure[\textit{House}. \label{fig:test_obj_house100}]{\includegraphics[width=0.32\textwidth]{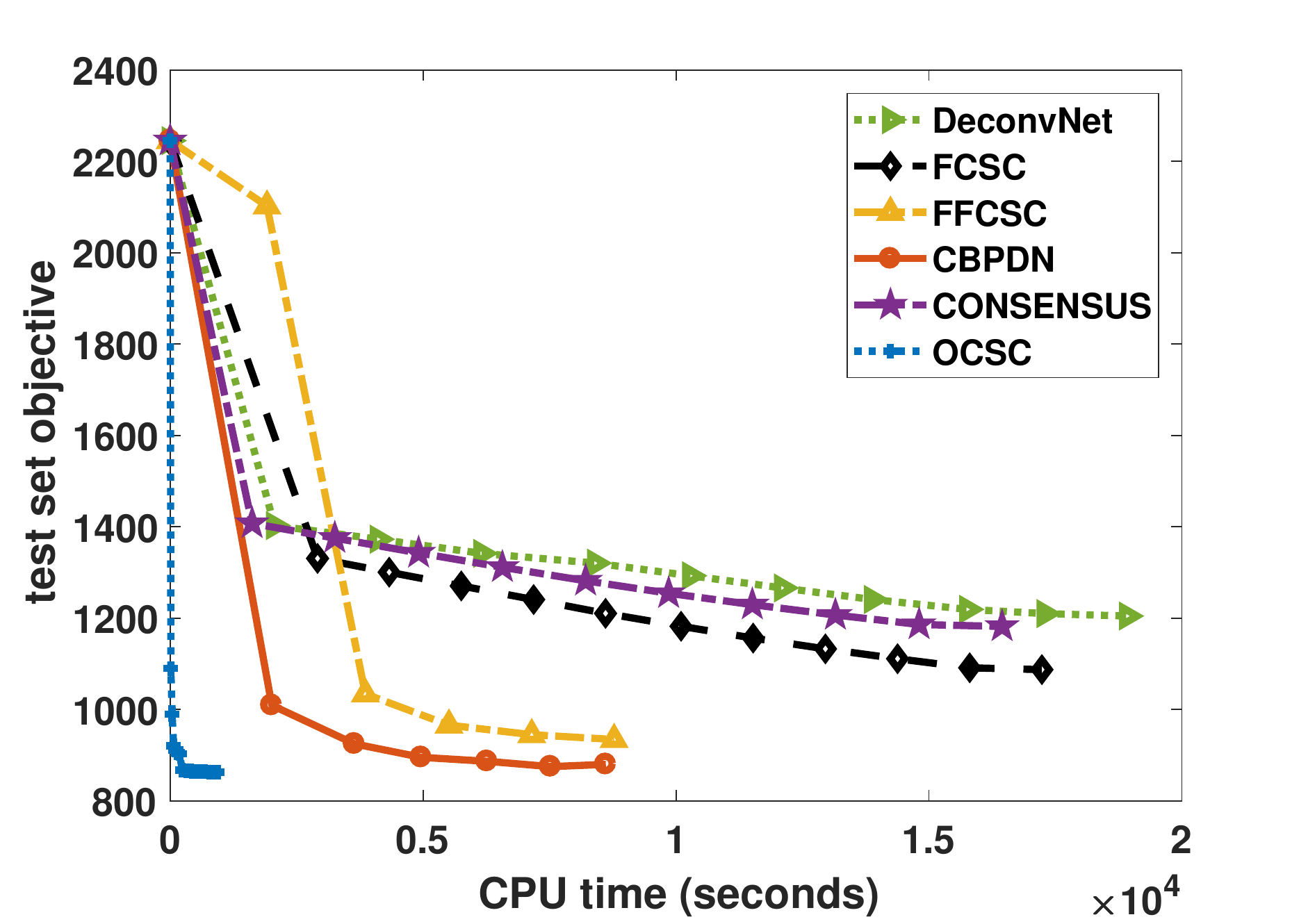}}	
	\caption{Test set objective vs CPU time on the small data sets.}
	\label{fig:test_time_small}
\end{figure}

In this section, we perform experiments on the small data sets (\textit{Fruit}, \textit{City} and
\textit{House}). 
Figure~\ref{fig:test_time_small} shows the test set objective vs CPU time. 
Though FFCSC 
has the same time complexity 
as OCSC,
OCSC is still much faster.
This is because
batch CSC methods 
update the dictionary
only after coding all the samples, 
while OCSC can refine the dictionary after coding each sample.
A similar behavior is also observed
between online and batch SC methods \cite{mairal2010online}.

Table~\ref{tab:eval_img_small} compares
the image reconstruction performance. 
As 
OCSC converges to a lower
objective than its batch counterparts (Figure~\ref{fig:test_time_small}),
it also outperforms the others
in terms of image reconstruction quality. 
Among all methods, OCSC performs the best, 
which is then followed by
CBPDN, FCSC and FFCSC.
CONSENSUS and DeconvNet perform the worst.

\begin{table}[ht]
	\caption{Testing PSNR obtained on the small data sets. 
		The highest and comparable PSNR (according to the pairwise t-test with 95\% confidence) are in bold.}
	\centering
	\begin{tabular}{c|c|c|c } \hline
		& \textit{Fruit} & \textit{City}&\textit{House}   \\\hline		
		DeconvNet &27.41$\pm$0.13 &28.11$\pm$0.25&25.39$\pm$0.27    \\\hline
		FCSC &27.90$\pm$0.18 &28.20$\pm$0.31&25.68$\pm$0.50   \\\hline
		FFCSC &28.13$\pm$0.15 &28.58$\pm$0.16&28.48$\pm$0.04  \\\hline	
		CBPDN &28.01$\pm$0.04&28.67$\pm$0.37&29.40$\pm$0.43   \\\hline
		CONSENSUS &27.62$\pm$0.14 &28.19$\pm$0.20&25.41$\pm$0.37   \\\hline		
		OCSC &\textbf{28.61$\pm$0.06}& \textbf{28.86$\pm$0.13}&\textbf{29.68$\pm$0.06}   \\\hline
	\end{tabular}
	\label{tab:eval_img_small}
\end{table}

\begin{figure}[ht]
	\centering
	{\includegraphics[width=0.13\textwidth]{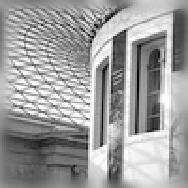}}
	{\includegraphics[width=0.13\textwidth]{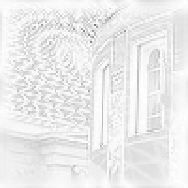}}
	{\includegraphics[width=0.13\textwidth]{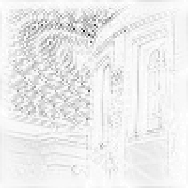}}
	{\includegraphics[width=0.13\textwidth]{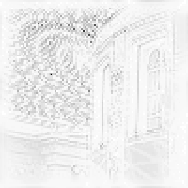}}
	{\includegraphics[width=0.13\textwidth]{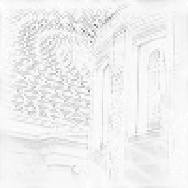}}	
	{\includegraphics[width=0.13\textwidth]{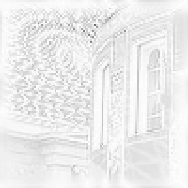}}
	{\includegraphics[width=0.13\textwidth]{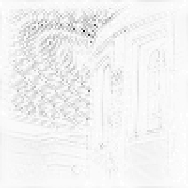}}	
	
	\vspace{5px}
	
	{\includegraphics[width=0.13\textwidth]{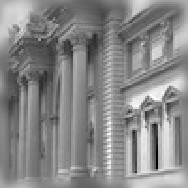}}
	{\includegraphics[width=0.13\textwidth]{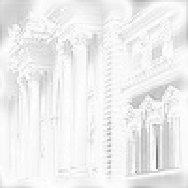}}
	{\includegraphics[width=0.13\textwidth]{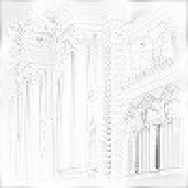}}
	{\includegraphics[width=0.13\textwidth]{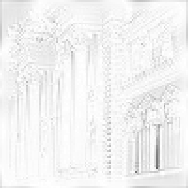}}
	{\includegraphics[width=0.13\textwidth]{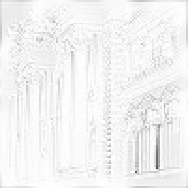}}	
	{\includegraphics[width=0.13\textwidth]{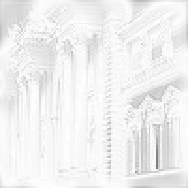}}
	{\includegraphics[width=0.13\textwidth]{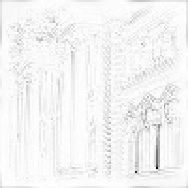}}	
	
	\vspace{5px}
	
	{\includegraphics[width=0.13\textwidth]{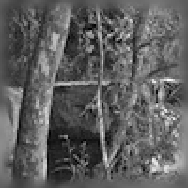}}
	{\includegraphics[width=0.13\textwidth]{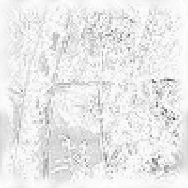}}
	{\includegraphics[width=0.13\textwidth]{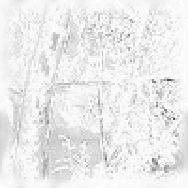}}
	{\includegraphics[width=0.13\textwidth]{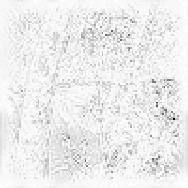}}
	{\includegraphics[width=0.13\textwidth]{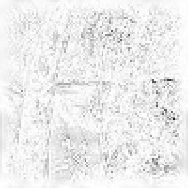}}	
	{\includegraphics[width=0.13\textwidth]{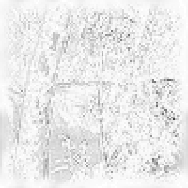}}
	{\includegraphics[width=0.13\textwidth]{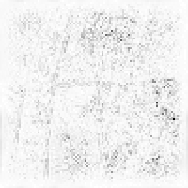}}			
	
	\subfigure	[Ground truth.\label{fig:clean}]
	{\includegraphics[width=0.13\textwidth]
		{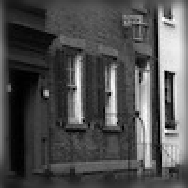}}
	\subfigure[DeconvNet.
	\label{fig:deconvnet_dif}]{\includegraphics[width=0.13\textwidth]
		{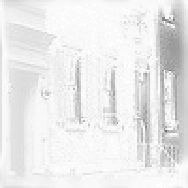}}
	\subfigure[FCSC.
	\label{fig:fcsc_dif}]{\includegraphics[width=0.13\textwidth]
		{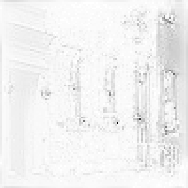}}
	\subfigure[FFCSC.
	\label{fig:ffcsc_dif}]{\includegraphics[width=0.13\textwidth]
		{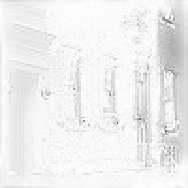}}
	\subfigure[CBPDN.
	\label{fig:cbpdn_dif}]{\includegraphics[width=0.13\textwidth]
		{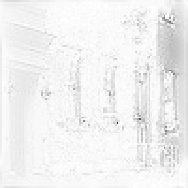}}	
	\subfigure[CONSENSUS.
	\label{fig:consensus_dif}]{\includegraphics[width=0.13\textwidth]
		{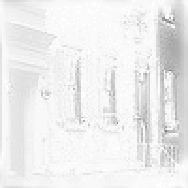}}
	\subfigure[OCSC.
	\label{fig:ocsc_dif}]{\includegraphics[width=0.13\textwidth]
		{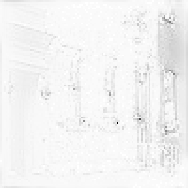}}			
	{\includegraphics[width = 0.6\textwidth]{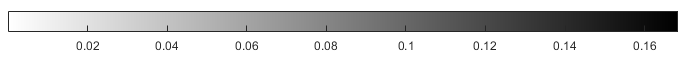}}		
	\caption{Differences between the ground truth and reconstruction on test set images
		of the \textit{Fruit} data set. }
	\label{fig:dif}
\end{figure}

Figure~\ref{fig:dif}  
shows the difference between the reconstructed images 
and the corresponding ground truths of some sample test set images
from the \textit{Fruit} data set.
As can be seen, the images reconstructed by OCSC are most similar to the ground truths,
which is then followed by CBPDN, FFCSC and FCSC. Images reconstructed by CONSENSUS and
DeconvNet are the least similar to the ground truths. A similar observation is also
observed on 
\textit{City} and \textit{House}, which
are not reported here because of the lack of space.

\begin{figure*}[bht]
	\centering
	\subfigure[DeconvNet.
	\label{fig:deconvnet_fruit}]{\includegraphics[width=0.15\textwidth]
		{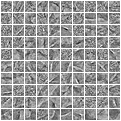}}
	\subfigure[FCSC.
	\label{fig:fcsc_fruit}]{\includegraphics[width=0.15\textwidth]
		{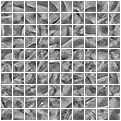}}
	\subfigure[FFCSC.
	\label{fig:ffcsc_fruit}]{\includegraphics[width=0.15\textwidth]
		{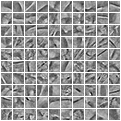}}
	\subfigure[CBPDN.
	\label{fig:cbpdn_fruit}]{\includegraphics[width=0.15\textwidth]
		{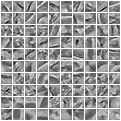}}	
	\subfigure[CONSENSUS.
	\label{fig:consensus_fruit}]{\includegraphics[width=0.15\textwidth]{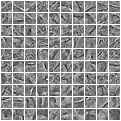}}
	\subfigure[OCSC.
	\label{fig:ocsc_fruit}]{\includegraphics[width=0.15\textwidth]{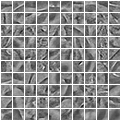}}	
	\caption{Dictionaries learned on the \textit{Fruit} data set.}
	\label{fig:dic_fruit}
\end{figure*}

Figure~\ref{fig:dic_fruit} shows the dictionaries learned on \textit{Fruit} (results on \textit{City} and \textit{House}
are similar).
As in \cite{aharon2006rm},
the filters are sorted in ascending order of the variance.\footnote{For
	filter $D(:,k)$, its variance is defined as $\frac{1}{M-1}\sum_{m=1}^{M}(D(m,k)-\frac{1}{M}\sum_{m=1}^{M}D(m,k))^2$.} As can be seen, dictionaries learned by FCSC, FFCSC, CBPDN and OCSC contain Gabor-style filters, while the dictionaries
learned by DeconvNet and CONSENSUS are vague.
We speculate that this is due to slow convergence of these two methods, as observed in 
Figure~\ref{fig:test_time_small}.

\begin{figure}[ht]
	\centering
	\subfigure[\textit{Flower}.  \label{fig:test_err_batch_flower}]
	{\includegraphics[width=0.32\textwidth]{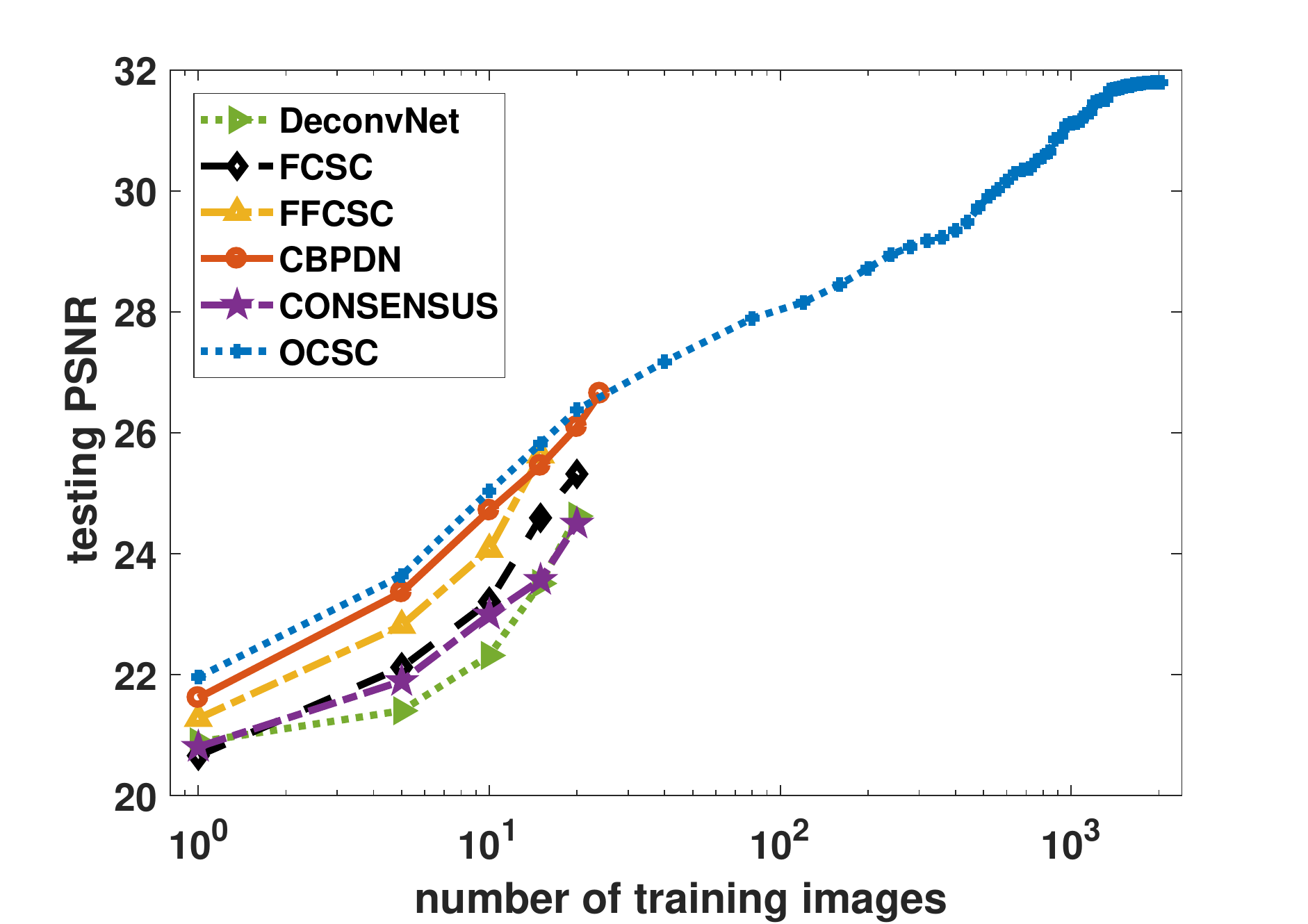}}
	\subfigure[\textit{Dog}.  \label{fig:test_err_batch_dog}]
	{\includegraphics[width=0.32\textwidth]{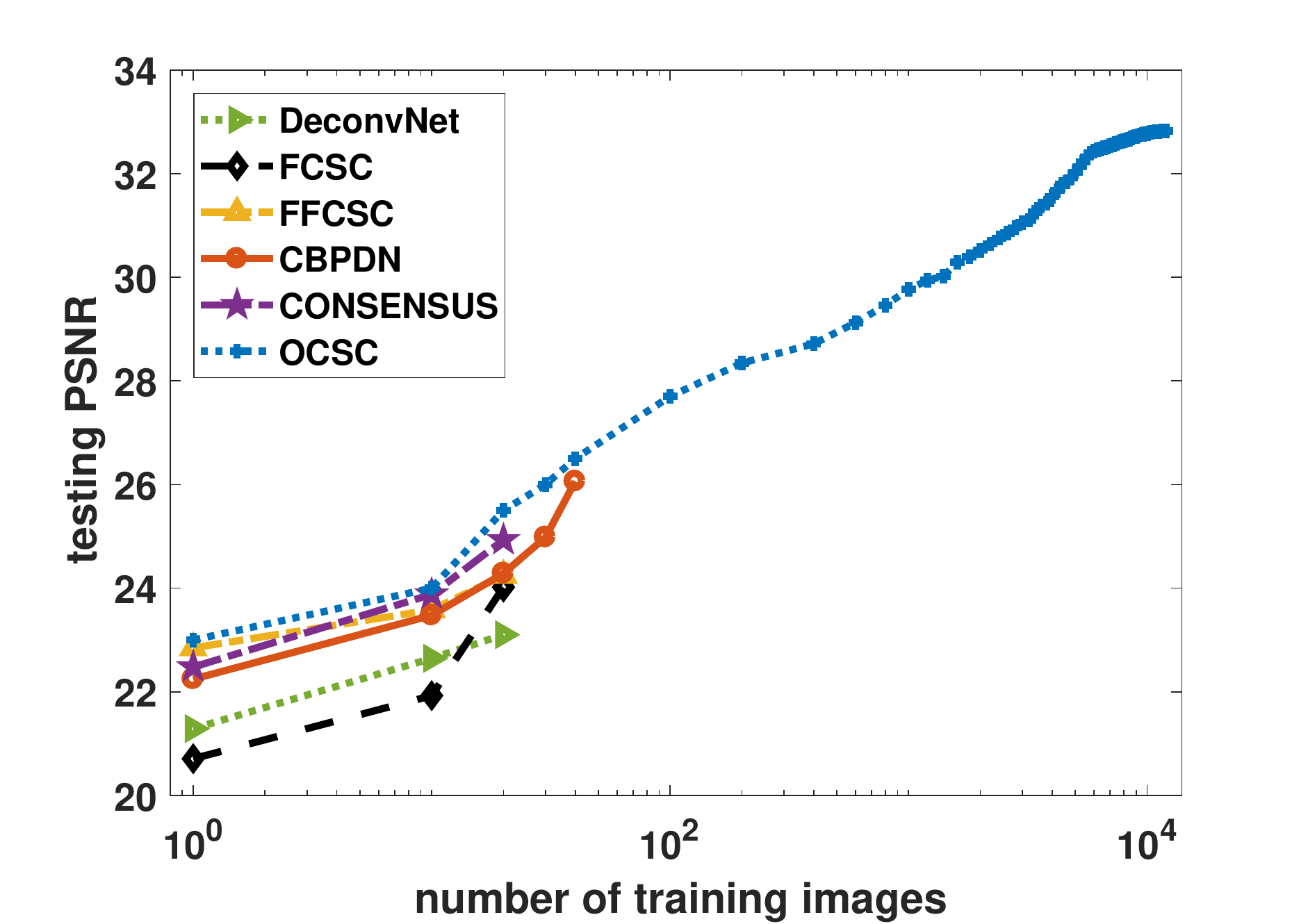}}
	\subfigure[\textit{CIFAR-10}.
	\label{fig:test_err_batch_c}]	{\includegraphics[width=0.32\textwidth]{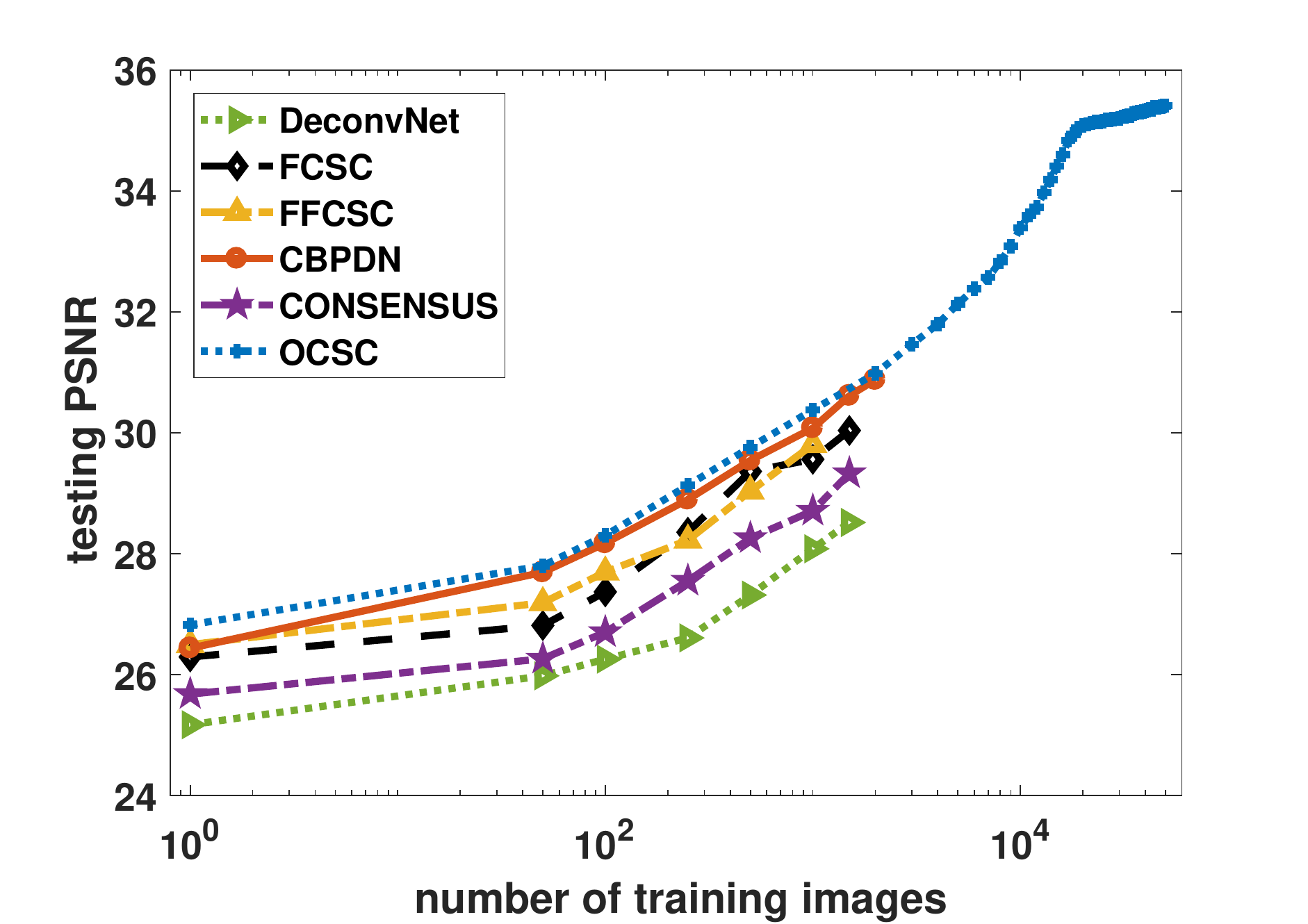}}
	\caption{Testing PSNR with varying number of training images.}
	\label{fig:test_large_err}
\end{figure}

\begin{figure}[ht]
	\centering
	\subfigure[\textit{Flower}.\label{fig:train_time_batch_flower}]
	{\includegraphics[width=0.32\textwidth]{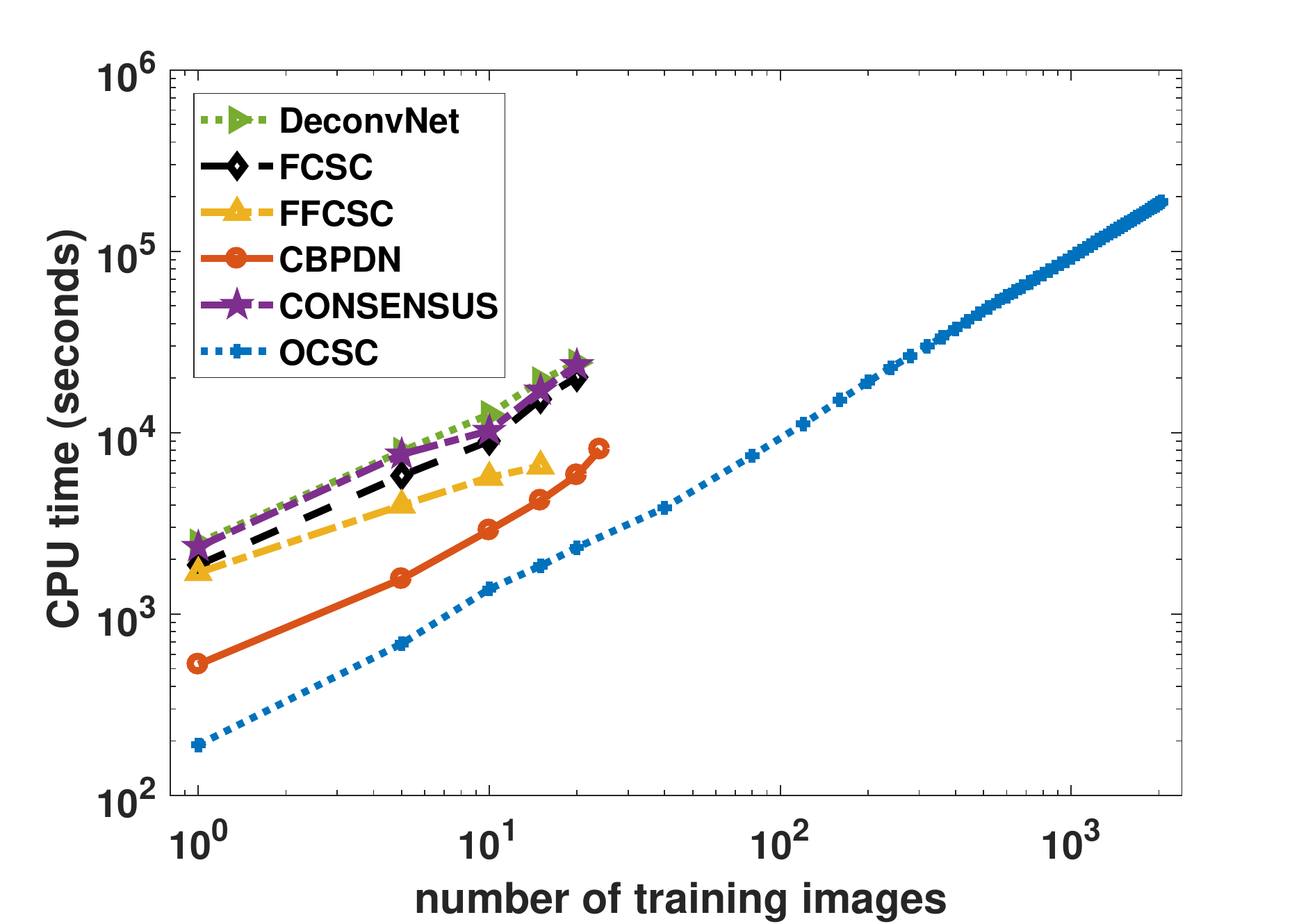}}
	\subfigure[\textit{Dog}.\label{fig:train_time_batch_dog}]
	{\includegraphics[width=0.32\textwidth]{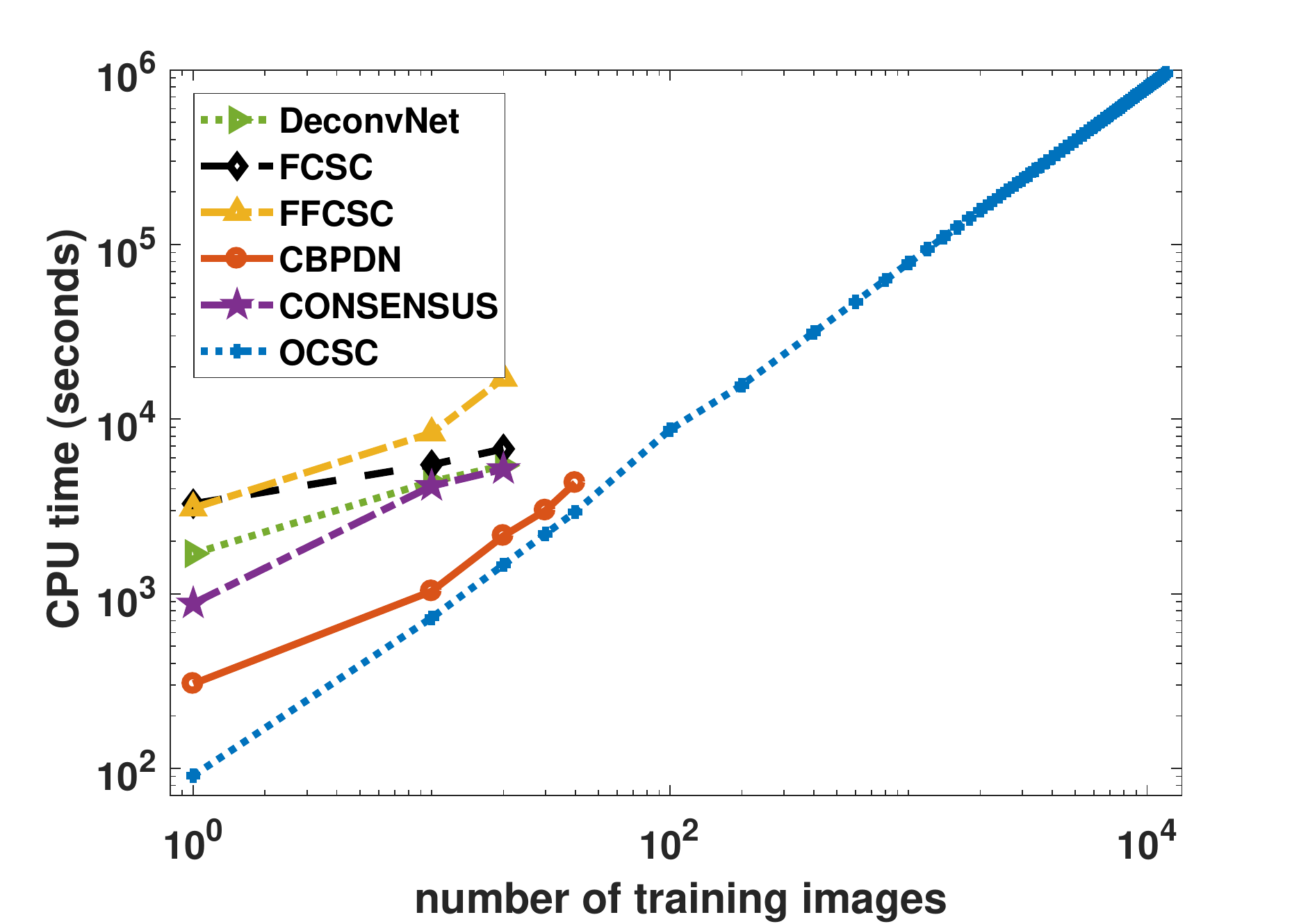}}
	\subfigure[\textit{CIFAR-10}.
	\label{fig:train_time_batch_cifar}]
	{\includegraphics[width=0.32\textwidth]{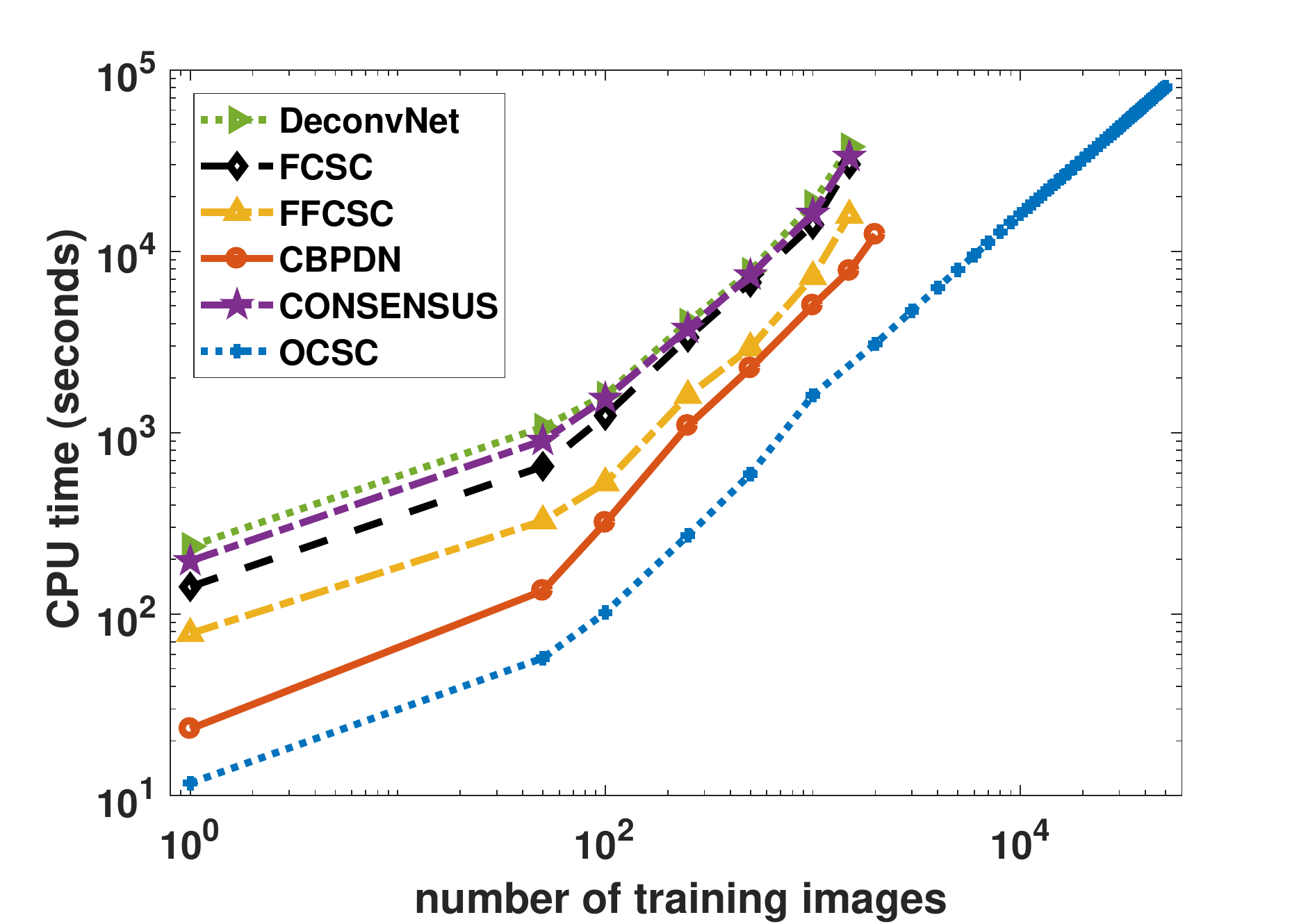}}		
	\caption{CPU time with varying number of training images.}
	\label{fig:train_large_time}
\end{figure}

\subsection{Large Data Sets}\label{sec:expts_large}

As discussed in Section~\ref{sec:complexity}, 
the space complexity of OCSC is independent of $N$. 
Hence, it 
can handle much larger training image sets than
batch CSC methods.
In this section, we illustrate this
on the \textit{Flower}, \textit{Dog}
and \textit{CIFAR-10} data sets.
The \textit{Flower} images are large ($500\times 500$), and the batch CSC methods quickly
run out of memory.\footnote{For $K=75$ or 100,
	FFCSC can only work with a single image, while
	the other batch methods can only handle
	2-5 images.}
To allow comparison with the batch methods,
we use $K$ as 50.

Figure~\ref{fig:test_err_batch_flower}
shows the testing PSNR with varying number of training images on the
\textit{Flower} data set.
As can be seen, performance is improved with more training images. However, the batch CSC
methods run out of memory quickly. Specifically,
CBPDN can only handle 25 training images,
while the other batch methods can only handle
20.
Figures~\ref{fig:test_err_batch_dog} and	\ref{fig:test_err_batch_c} show
results on \textit{Dog} and \textit{CIFAR-10}, respectively. 
Among the batch methods, 
CBPDN can handle 40 training images on \textit{Dog} and 2000 training images on
\textit{CIFAR-10}, while
the others can only handle 20 training images on \textit{Dog} and 1500 training images on \textit{CIFAR-10}.
On the other hand, OCSC can always handle the full training data set.

Figure~\ref{fig:train_large_time} shows the CPU time with different numbers of training
images.
Given the same amount of time, OCSC can train on more images than the batch methods. Combined with Figure~\ref{fig:test_large_err}, OCSC obtains higher testing PSNR than batch
methods given the same amount of time.

\section{Conclusion}

In this paper, we proposed a scalable convolutional sparse coding methods in the online setting. 
By reformulating the CSC objective,
the sizes of the history matrices required in the online setup 
can be significantly reduced.
Moreover, the resultant optimization problem can be efficiently solved using ADMM, with
closed-form solutions for the ADMM subproblems.
We also provide theoretical guarantee on the convergence of the  proposed algorithm.
While the batch CSC methods need large space and heavy computation cost, 
extensive experiments show that
the proposed
method can efficiently learn from large data sets with much less space. 

As for future work, we will consider introducing nonconvex regularizers to CSC. Traditionally,
the convex $\ell_1$-regularizer is used to encourage sparsity among the codes. Though this
leads to easier optimization, the resultant code may not be as sparse and accurate as when a
nonconvex regularizer
(such as the log-sum-penalty) is used \cite{candes2008enhancing}.  
Moreover,
on large data sets,
many filters are often needed to capture the presence of more local patterns.
Though the proposed online algorithm has reduced the space consumption significantly
compared to batch CSC methods, its space complexity still depends on the number of filters.
A promising direction
is to further reduce the storage by approximating the filters as
linear combinations of a few base filters.

\bibliographystyle{IEEEtran}
\bibliography{csc}

\appendices

\section{Proofs}

\subsection{Proposition~\ref{pr:reorder}}
\label{app:reorder}

\begin{proof}
	\eqref{eq:tmp1} can be expanded as
	\begin{align*}
	\arg\min_{\tilde{D}}
	\frac{1}{2tP}\sum_{i=1}^{t}\left\|\tilde{x}_i-\dot{D}\dot{z}_i\right\|^2_2 
	=\arg\min_{\tilde{D}} \frac{1}{2P}\tr\left(\dot{D}^{\dagger}\dot{D}
	\left( \sum_{i=1}^{t}\frac{1}{t}\dot{z}_i\dot{z}_i^{\dagger}\right) -2\dot{D}^{\dagger}\left( \sum_{i=1}^{t}\frac{1}{t}\tilde{x}_i\dot{z}_i^{\dagger}\right) \right),
	\end{align*} 
	where
	$\dot{D}=[\text{Diag}(\tilde{D}(:,1)),\dots,\text{Diag}(\tilde{D}(:,K))]$, and
	$\dot{z}_i=\text{vec}(\tilde{Z}_i)$.
	Thus,
	\begin{eqnarray*}
		\dot{D}^{\dagger}\dot{D}
		\!\!\!&= &\!\!\!
		\left[
		\begin{array}{c}
			\begin{smallmatrix}
				\tilde{D}(1,1) & & 0 \\
				&\ddots&\\
				0 & & \tilde{D}(P,1)\rule[-1ex]{0pt}{2ex}
			\end{smallmatrix} \\
			\ddots
			\\
			\begin{smallmatrix}\rule{0pt}{2ex}
				\tilde{D}(1,K) & & 0 \\
				&\ddots&\\
				0 & & \tilde{D}(P,K)
			\end{smallmatrix}  
		\end{array} 
		\right]
		\cdot
		\left[
		\begin{array}{ccc}
			\begin{smallmatrix}
				\tilde{D}(1,1) & & 0 \\
				&\ddots&\\
				0 & & \tilde{D}(P,1)\rule[-1ex]{0pt}{2ex}
			\end{smallmatrix}
			\ddots
			\begin{smallmatrix}\rule{0pt}{2ex}
				\tilde{D}(1,K) & & 0 \\
				&\ddots&\\
				0 & & \tilde{D}(P,K)
			\end{smallmatrix}  
		\end{array} 
		\right].
	\end{eqnarray*}		
	For $\tilde{D}(j,k)$, only when it is multiplied by elements of the same row index $\tilde{D}(j,l)$, $l=1,\dots,K$, will result in non-zero values.
	In other words, multiplications among different rows 
	can be avoided. 
	We then
	directly operate on each row of $\dot{D}$, which is exactly $\tilde{D}(p,:)$ if zeros are dropped. Consequently, we have the form in Proposition~\ref{pr:reorder}.
\end{proof}

\subsection{Lemma~\ref{lem:bndD}}
\label{app:lemD}
\begin{proof}
	Obviously, 
	$\mathcal{S}$ is a convex constraint set.
	To show  $\mathcal{S} \subset \mathcal{D}$, 
	recall that $\bar{D}=[T(D(:,1)),\dots,T(D(:,K))]$, where each $T(D(:,k))$ is the Toeplitz matrix associated with $D_t(:,k)$ as defined in \eqref{eq:csc2sc}.
	Because of the constraint $\|D(:,k)\|_2^2\le1$,
	each column $\bar{D}(:,l)$ of $\bar{D}$, 
	satisfies $\|\bar{D}(:,l)\|_2^2\le 1$. 
	Thus $\|\bar{D}\|_F^2\le KP$.
	Hence,
	$\mathcal{S} \subset \mathcal{D}$ and $\mathcal{S}$ is bounded.
\end{proof}

\section{Experimental Comparison with \cite{degraux2017online,liu2017online}}
\label{sec:extraExp}

In this section,
we provide extra experimental results comparing OCSC with concurrent works
OCDL1 \cite{degraux2017online} and OCDL2 \cite{liu2017online} discussed in Section~\ref{sec:discuss}.
We use implementation by ourself as no pubic codes are available.

\subsection{Large Data Sets}
We experiment on large data sets to test these online methods' scalability.
The evaluation metric used on empirical convergence is test set objective, on image reconstruction quality is PSNR.
\textit{CIFAR-10} is used, which has a large number of small images ($P=32\times32$).
This is because a single image of \textit{Flower} ($P=500\times500$) makes both OCDL1 and OCDL2 run out of memory, even \textit{Dog} ($P=224\times224$) makes OCDL2 not work.
Hence for complete comparison, we use \textit{CIFAR-10} only.

We still use the same training and test splits as in Section~\ref{sec:expt}. 
Hyper-parameters are the same as Section~\ref{sec:expt}.
As a recap, we set $\beta=1$, $M=11$ and
$K=100$. 
The methods are stopped when the
relative changes in $Z_t$ (i.e.,
$\|Z_t-Z_t^{\text{old}}\|_2/\|Z_t\|_2$) and
$D$ (i.e.,
$\|D-D^{\text{old}}\|_2/\|D\|_2$) are both smaller than ${10}^{-3}$.

\begin{figure}[ht]
\centering
\subfigure[CPU time v.s testing objective.\label{fig:con_obj}]
{\includegraphics[width=0.32\textwidth]{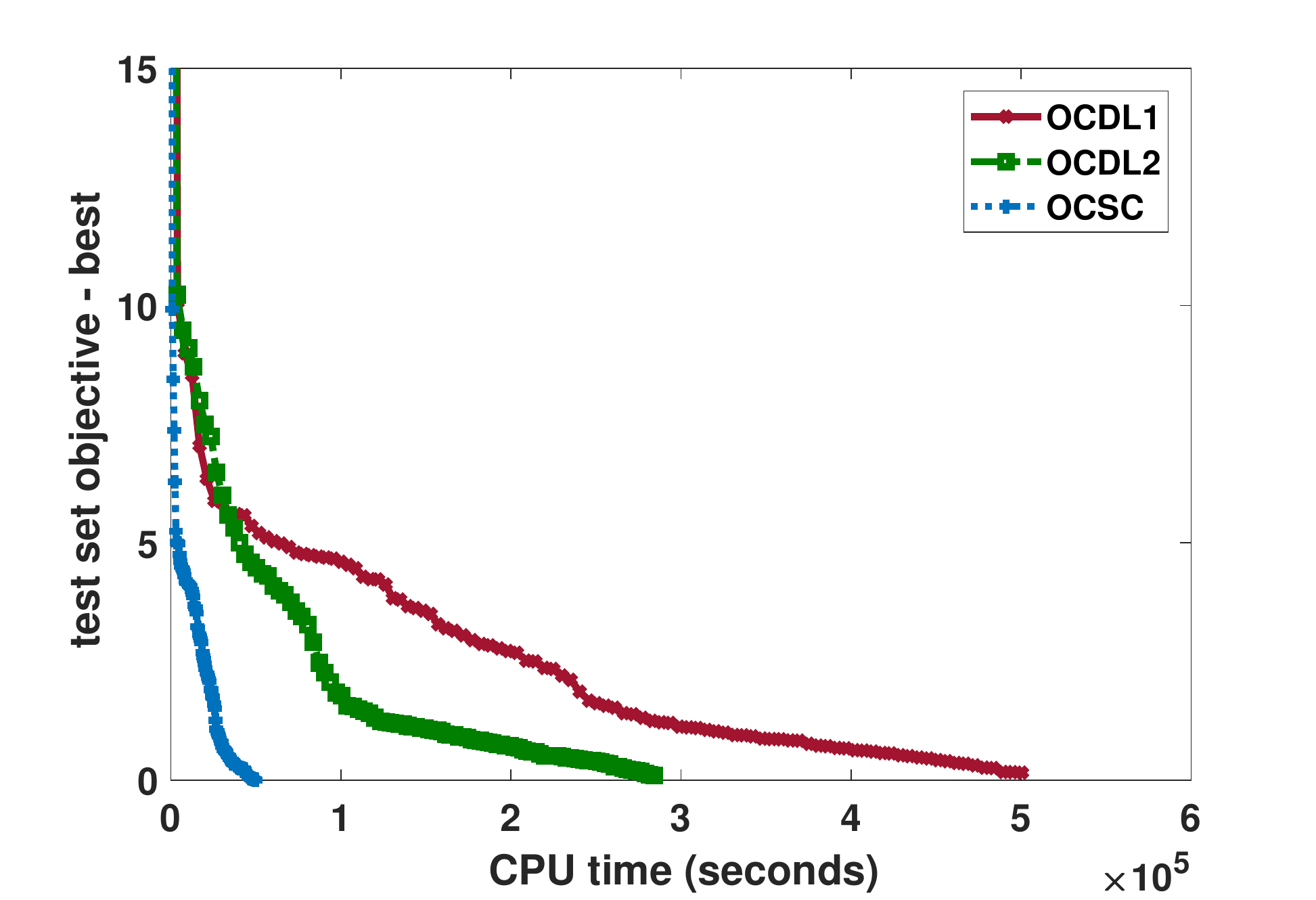}}
\subfigure[CPU time v.s testing PSNR.\label{fig:con_psnr}]
{\includegraphics[width=0.32\textwidth]{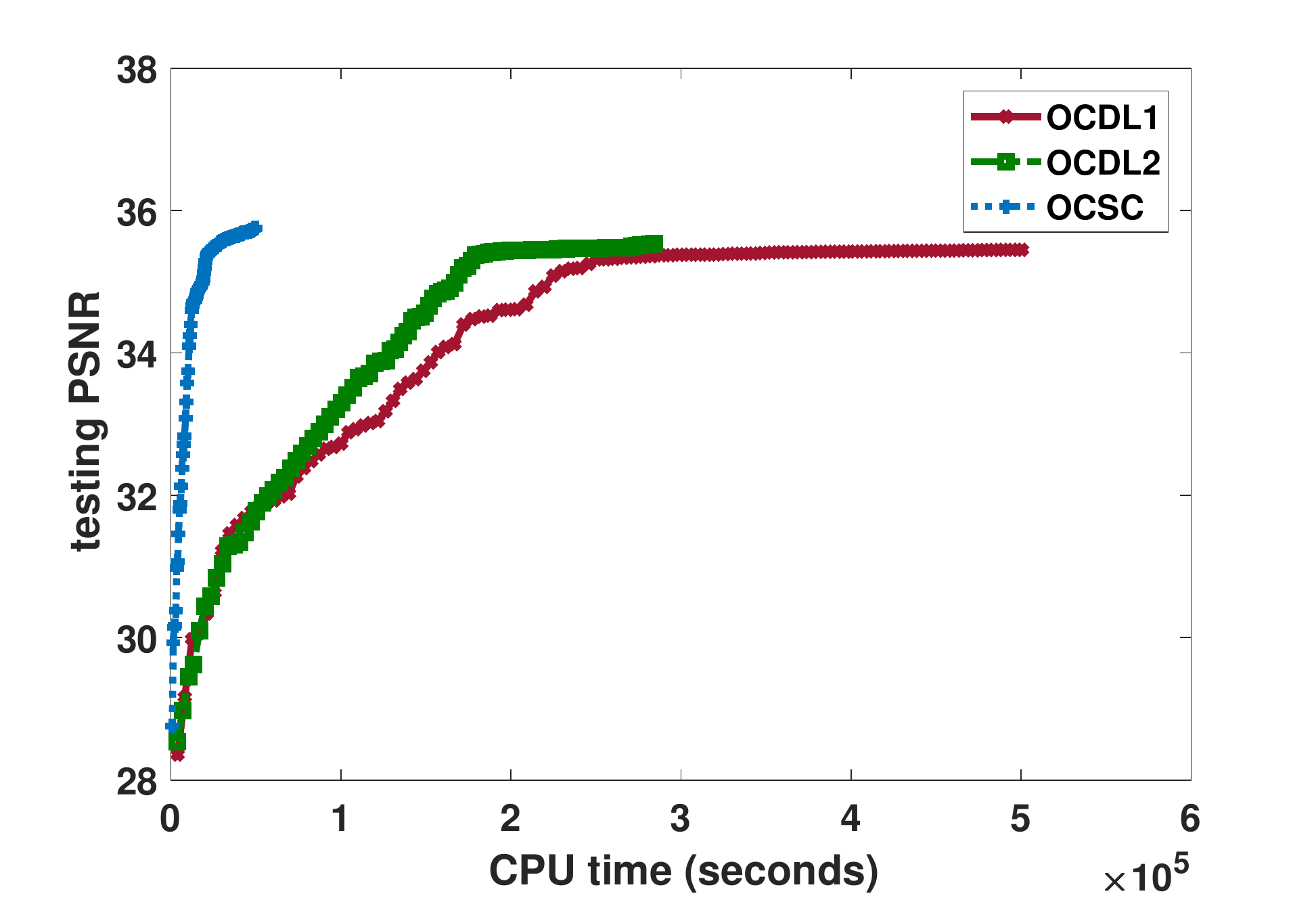}}
\caption{Comparison with OCDL1 \cite{degraux2017online} and OCDL2 \cite{liu2017online}
	on \textit{CIFAR-10} dataset.}
\vspace{-10px}
\end{figure}

Figure~\ref{fig:con_obj} shows the test set objective vs CPU time.
As shown, 
OCSC converges the fastest, 
then OCDL2 follows, OCDL1 is the slowest. 
OCDL1 lags behinds due to its slow convolution performed in spatial domain and the costly updating of history matrices.
As for OCDL2, although it shares the efficiency of performing convolution in frequency domain, it is very slow to deal with sparse matrices. 
They finally reach to comparable testing objective value as they are solving the same online CSC problem. 
Figure~\ref{fig:con_psnr}
shows the testing PSNR vs CPU time.
For all three methods, the testing PSNR is improved given more training images, and they can reach similar final testing PSNR.
However, the CPU time needed to reach it varies a lot. In other words, given same amount of time, OCSC always gets a higher testing PSNR.

\subsection{Synthetic Data}
\label{sec:appadmmfis}

In this section, we perform experiments on synthetic data to compare the convergence speed of ADMM (used by our OCSC) and FISTA (used by both OCDL1 \cite{degraux2017online} and OCDL2 \cite{liu2017online}) on CSC problem.
We evaluate the convergence speed by training set objective vs time.
We choose frequency domain to compare the efficiency of ADMM and FISTA.
As shown in Figure~\ref{fig:con_obj} and Figure~\ref{fig:con_psnr}, although OCDL1 and OCDL2 both use FISTA, OCDL1 which operates in spatial domain is much slower than OCDL2 which operates in frequency domain.
Specifically,
we compare proposed ADMM on \eqref{eq:ocsc_dic} with 
FISTA on \eqref{eq:ocdl2} \cite{liu2017online}.
Note that the two problems are equivalent to \eqref{eq:ocsc},
thus we compute the objective based on \eqref{eq:ocsc}.


We set $K = 100$,
and the size of synthetic data $x \in \R^{P}$ where $P=10000$.
The entries of $x$, $D$ and $Z$ are sampled i.i.d. from the standard normal distribution $\mathcal{N}(0,1)$,
and each $D(:,:,k), k=1\dots K$ is projected to the $\ell_2$-norm unit ball.

\begin{figure}[H]
\centering
\subfigure[Objective v.s number of iterations.\label{fig:admmfist_obj}]
{\includegraphics[width=0.32\textwidth]{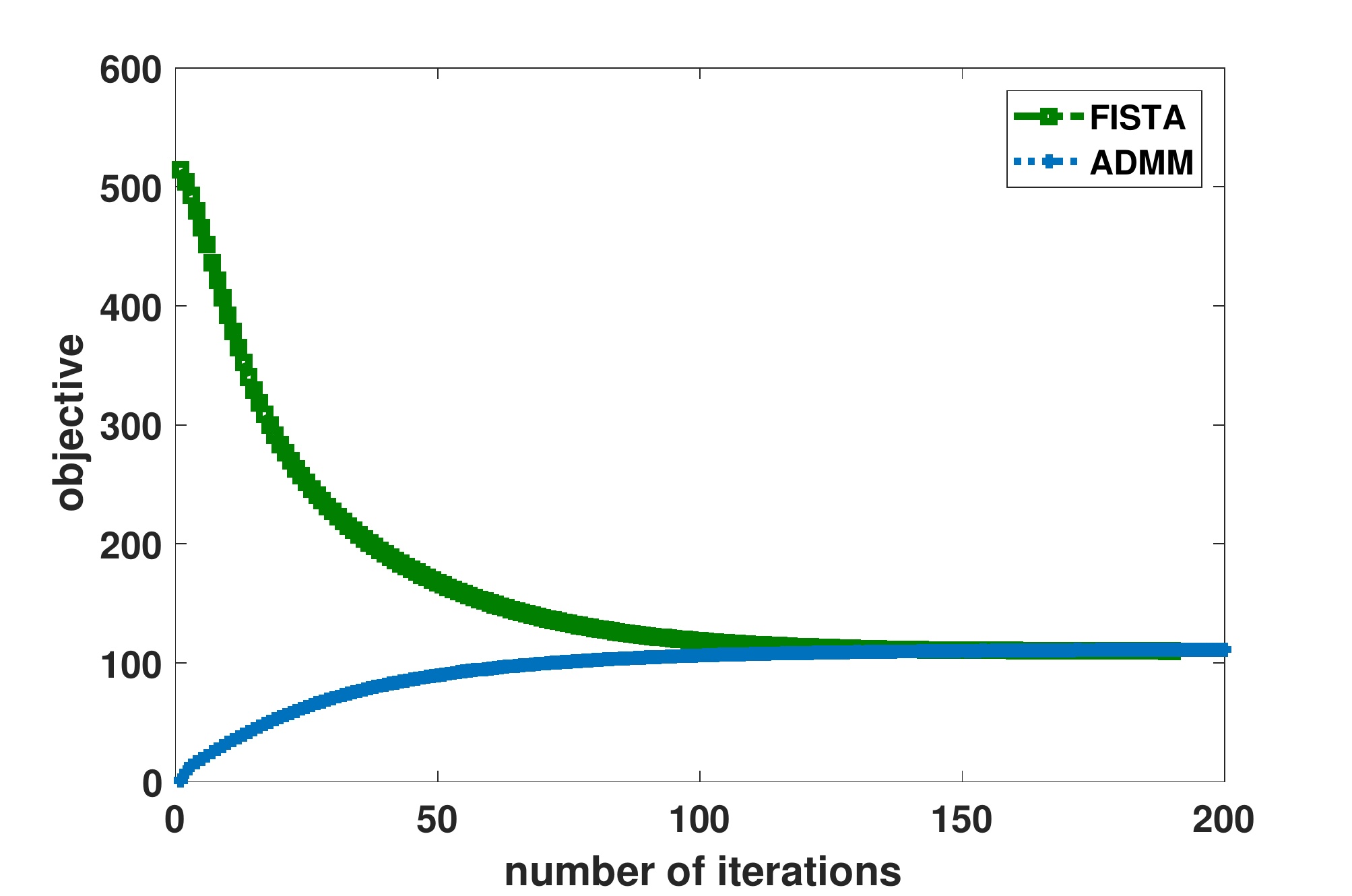}}
\subfigure[Objective v.s CPU time (seconds).\label{fig:admmfist_time}]
{\includegraphics[width=0.32\textwidth]{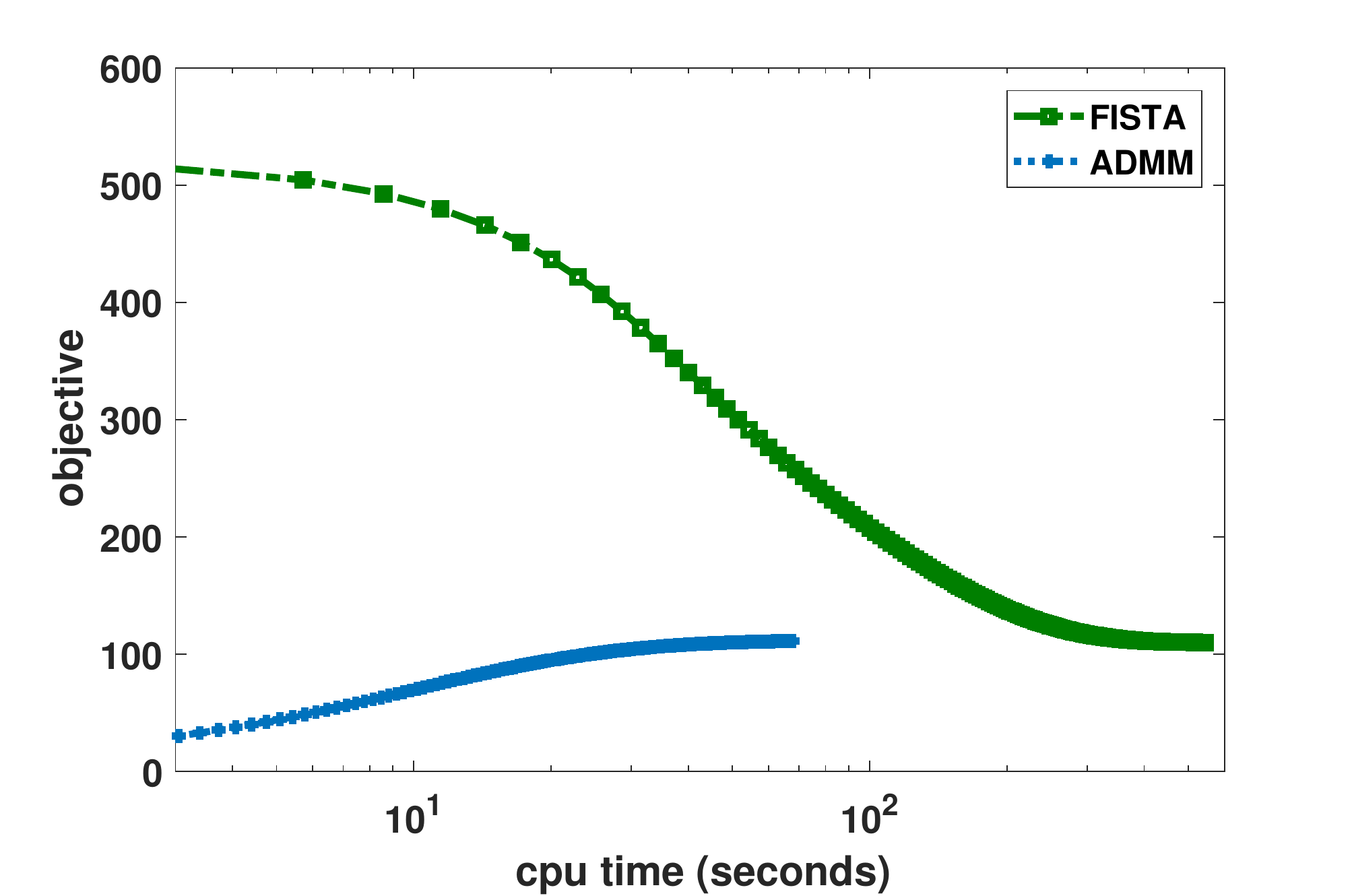}}
\caption{Comparison of FISTA and ADMM on the synthetic data.}
\label{fig:admmfist}
\end{figure}

The comparison of objective vs iterations and CPU time are shown in Figure~\ref{fig:admmfist_obj} and Figure~\ref{fig:admmfist_time}.
In terms of the number of iterations, ADMM converges faster than FISTA.
Then, in terms of time, 
FISTA is much slower than ADMM. 
This relates to the design of OCDL2 mentioned in \cite{liu2017online}.
Empirically, for our data, OCDL2's history matrices need hundreds of GBs in dense format.
Hence to fit into memory, OCDL2 has to operate on sparse matrices, which is much slower on matrix multiplications.
This also verifies our discussion in Section~\ref{sec:discuss}, ADMM converges faster than FISTA on CSC problem.

\end{document}